\pdfoutput=1

\documentclass[11pt]{article}

\usepackage{authblk} 

\usepackage{EMNLP2023}

\usepackage{times}
\usepackage{latexsym}

\usepackage[T1]{fontenc}

\usepackage[utf8]{inputenc}

\usepackage{microtype}

\usepackage{inconsolata}

\usepackage{graphicx} 
\usepackage{subfigure} 
\usepackage{color} 
\usepackage{arydshln} 
\usepackage{amsmath, amsfonts, amssymb, bm} 
\usepackage{float} 
\usepackage{placeins} 
\usepackage{multirow} 
\usepackage{xcolor} 
\usepackage{enumitem} 
\usepackage{tablefootnote} 
\usepackage{longtable}
\usepackage[normalem]{ulem} 
\usepackage{algorithm} 
\usepackage{algorithmic} 

\DeclareMathAlphabet{\mathbbmsl}{U}{bbm}{m}{sl}

\newcommand{\ColorUp}[1]{\textcolor{purple}{$\uparrow$#1}\ \ }
\newcommand{\ColorDown}[1]{\textcolor{teal}{$\downarrow$#1}\ \ }
\newcommand{\ColorUpCR}[1]{\textcolor{teal}{$\uparrow$#1}\ \ }
\newcommand{\ColorDownCR}[1]{\textcolor{purple}{$\downarrow$#1}\ \ }
\newcommand{\ColorBest}[1]{\textcolor{orange}{#1}}

\newcommand\scalemath[2]{\scalebox{#1}{\mbox{\ensuremath{\displaystyle #2}}}}

\definecolor{lightred2}{HTML}{ea9999}
\definecolor{lightyellow2}{HTML}{ffe599}
\definecolor{lightblue2}{HTML}{9fc5e8}
\definecolor{lightgreen2}{HTML}{b6d7a8}

\definecolor{lightred1}{HTML}{e06666}
\definecolor{lightyellow1}{HTML}{ffd966}
\definecolor{lightblue1}{HTML}{6fa8dc}
\definecolor{lightgreen1}{HTML}{93c47d}

\title{Conceptor-Aided Debiasing of Large Language Models}

\author{
    Li S. Yifei$^1$, Lyle Ungar$^1$, João Sedoc$^2$,
    \vspace{0.01in} \\
    $^1$University of Pennsylvania, $^2$New York University\\
    \texttt{\{liyifei, ungar\}@upenn.edu, jsedoc@stern.nyu.edu} 
}

\begin{document}
\setlength{\abovedisplayskip}{3pt}
\setlength{\belowdisplayskip}{3pt}
\maketitle
\begin{abstract}
Pre-trained large language models (LLMs) reflect the inherent social biases of their training corpus. Many methods have been proposed to mitigate this issue, but they often fail to debias or they sacrifice model accuracy.  We use {\it conceptors}--a soft projection method--to identify and remove the bias subspace in LLMs such as BERT and GPT. We propose two methods of applying conceptors (1) bias subspace projection by post-processing by the conceptor NOT operation; and (2) a new architecture, conceptor-intervened BERT (CI-BERT), which explicitly incorporates the conceptor projection into all layers during training. We find that conceptor post-processing achieves state-of-the-art (SoTA) debiasing results while maintaining LLMs' performance on the GLUE benchmark. Further, it is robust in various scenarios and can mitigate intersectional bias efficiently by its AND operation on the existing bias subspaces. Although CI-BERT's training takes all layers' bias into account and can beat its post-processing counterpart in bias mitigation, CI-BERT reduces the language model accuracy. We also show the importance of carefully constructing the bias subspace. The best results are obtained by removing outliers from the list of biased words, combining them (via the OR operation), and computing their embeddings using the sentences from a cleaner corpus.\footnote{Code link: \url{https://github.com/realliyifei/conceptor-debias-llm}.}
\end{abstract}

\section{Introduction}
LLMs such as BERT~\cite{devlin2019bert} and GPT \citep{radford2019language,brown2020language} are extremely successful in most natural language processing (NLP) tasks. However, since they are trained on texts written by humans, the social bias is inherited and represented in the parameters of LLMs~\citep{bolukbasi2016man,Caliskan2022GenderBI}.
For example, gender bias has been found in contextualized embeddings (\citealp{may2019seat}; \citealp{zhao2019gender}). Therefore, many researchers have developed debiasing techniques to improve the social fairness of NLP. However, such debiasing often fails to debias effectively and reduces language model performance in downstream tasks~\cite{meade_2022_empirical}. Furthermore, most debiasing methods neither follow ~\citet{bommasani2020interpreting}'s suggestion to reduce bias in all layers nor tackle intersectional bias in an efficient way~\cite{lalor2022intersectional}.

\begin{figure}[t!]
    \centering
    \includegraphics[scale=0.25]{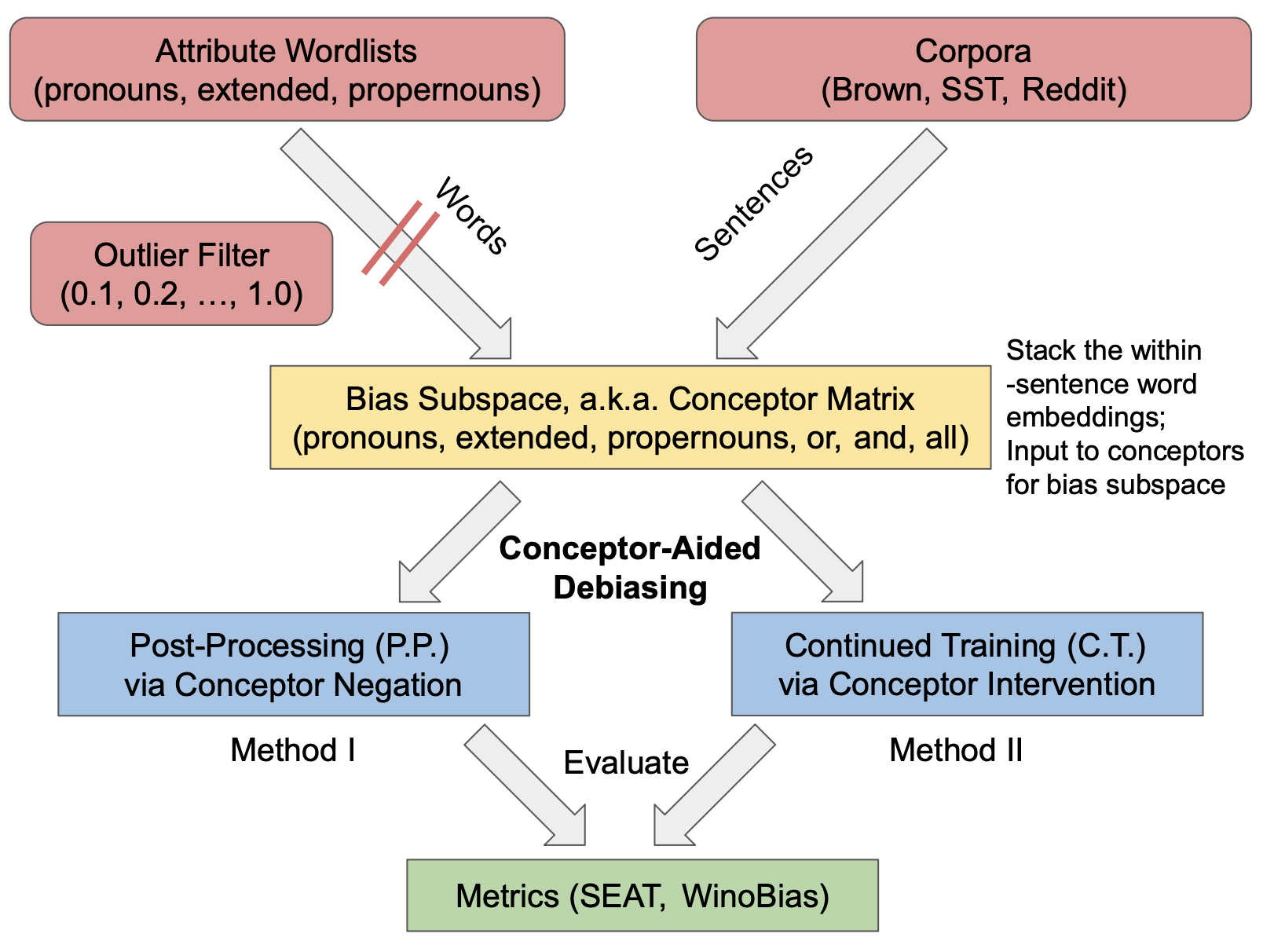}
    \caption{The pipeline of the conceptor-aided debiasing paradigm. We first use different \textcolor{lightred1}{settings (wordlists with outlier filter and corpora)} to generate the best \textcolor{lightyellow1}{bias subspace (conceptor matrix)}, then apply them to two \textcolor{lightblue1}{conceptor-aided debiasing methods} and measure the debiasing performance by two \textcolor{lightgreen1}{evaluation metrics}. The experiment is conducted on two LLMs: BERT and GPT.}
    \label{fig:pipeline}
\end{figure}

In this paper, we challenge \citet{karve2019conceptor}'s conclusion that \textit{conceptor negation} fails to debias BERT stably. Instead, we are the first ones to empirically find that as a soft shrinkage of the principal components of the subspace defined by the list of biased words~\cite{liu2018correcting}, conceptors is a powerful tool to debias LLMs such as BERT and GPT using either post-processing or continued-training.
In this process, we demonstrate the effect on debiasing performance of choosing different corpora, subspace removal methods, and criteria for selecting the list of bias attribute words that are used to construct the bias subspace. 
Further, we unprecedentedly show that the conceptor can tackle varied types of biases (e.g. gender, race) intersectionally and efficiently by its unique logical operation. 

Specifically, the \textit{attribute wordlists} at the core of our method, and the methods we build on, are sets of attribute words related to bias. These typically come in opposing pairs (e.g. `man'/`woman', `prince'/`princess'). \citet{bolukbasi2016man}, \citet{liang2020towards} and others use the first principal component (PC) to define the \textit{bias subspace}--which can be later subtracted entirely to debias. We similarly construct such subspaces, but use conceptors as a `soft' way to remove them--downscale the PC adjusted by a regularized identity map. When generating such wordlists, it may be more representative of bias by removing outliers in the embedding space. Considering the embeddings are contextualized, we select the contextualized token-level word embeddings using sentences from a specific \textit{corpus}. Then we stack them to generate a bias subspace in a form of a conceptor matrix for the debiasing in the next step. The pipeline is shown in Figure~\ref{fig:pipeline}.

This work contributes the following:
\begin{itemize}[noitemsep,topsep=0pt]
\item Employs \textit{conceptor negation post-processing} to debias LLMs such as BERT and GPT, beating most SoTA while retaining useful semantics and robustness in multiple scenarios
\item Explores \textit{conceptor-intervened BERT (CI-BERT)}--a novel model architecture that continues training BERT after incorporating conceptors within all of BERT's layers
\item Illustrates how different corpora, bias attribute wordlists, and outlier removal criteria impact debiasing performance 
\item Demonstrates conceptor-aided methods can be generalized to different layers of LLMs and various types of biases and can mitigate them intersectionally by its unique logical operation
\end{itemize}

\section{Related Work} 


\subsection{Bias Manifestation} 

Multiple demographic biases are common in society. Among them, gender bias is the most well-studied in academia, given its omnipresence and bi-polarity (\citealp{bolukbasi2016man}; \citealp{may2019seat}; \citealp{kurita-etal-2019-measuring}). Other social biases (e.g. racial, religious) are also widespread in LLMs and attracting increasing attention \citep{nangia-etal-2020-crows,nadeem-etal-2021-stereoset,meade_2022_empirical}.

Such social bias manifests itself in all layers of the contextualized representations of LLMs like BERT and GPT~\cite{bommasani2020interpreting}; and \citet{kaneko2021debiasing} show that debiasing all layers is more effective. Moreover, \citet{lalor2022intersectional} indicates the importance of addressing varied biases in different dimensions. Thus, a new challenge is raised on how to adapt current methods or develop novel paradigms to mitigate the bias in each layer and across multiple social dimensions.

\subsection{Debiasing Techniques and Challenges}

We collect the mainstream SoTA debiasing methods (Overview: \citet{meade_2022_empirical,xie2023empirical}), each with typical examples: 

(1) Bias Subspace Projection (BSP): the classic method of bias subspace subtraction is to first capture the bias subspace determined by attribute words in the corpora and then project the bias direction out from the language embeddings. This can be done by post-processing as either hard projection (\citealp{bolukbasi2016man}; \textsc{\textbf{SentenceDebias}}, \citealp{liang2020towards}) or soft projection (\citealp{karve2019conceptor}). Some variants attain a similar goal by training a linear classifier (\textbf{INLP}, \citealp{ravfogel-etal-2020-null}) or fine-tuning LLMs~\cite{kaneko2021debiasing}. 

(2) Counterfactual Data Augmentation (\textbf{CDA}): swapping the bi-polar bias attribute words (e.g. her/him) to rebalance the training dataset and therefore decrease the gender bias (\citealp{webster2020measuring}; \citealp{barikeri-etal-2021-redditbias}). 

(3) Dropout Regularization (\textsc{\textbf{Dropout}}): in combination with an additional pre-training, increasing the dropout components inside the transformer-based language models can lead to lower bias~\citep{webster2020measuring}. 

(4) \textsc{Self-Debias}: by using specific templates to encourage LLMs to generate toxic output and then modifying the original output distribution of the model by a decoding algorithm, \citet{schick2021self} makes use of the internal knowledge of language model to debias in a post-hoc manner.

Further, it is common to combine multiple such methods. For instance, \citet{zhao2019gender} and \citet{liang2020towards} combine the techniques of data augmentation and hard debiasing.
However, per the discussion in~\citet{meade_2022_empirical}, the methods often neither debias as well as they claim (e.g. CDA, \textsc{Dropout}, \textsc{\textsc{SentenceDebias}}), nor do they maintain the model's capability for downstream tasks (e.g. CDA, \textsc{Dropout}, INLP). Worse, some techniques like CDA and \textsc{Dropout} increase the bias measured on SEAT--a test of language bias which we will describe in Section~\ref{sec:evaluate-bias}.

This dilemma challenges us to develop new methods to further reduce bias  while retaining meaningful semantics.  Last, the majority of debiasing methods ground the bias by word list; different lists can lead to different debias performance~\cite{mimno2021bad}.  

\subsection{Conceptors in NLP}\label{subsec:conceptors}

Conceptors--a soft projection method supporting conceptual abstraction and logical operations~\cite{jaeger2014controlling}--has been adapted to NLP domains such as debiasing \citep{liu2018correcting,sedoc2019role,karve2019conceptor}, continual learning~\cite{liu2019continual}, and semantic information enrichment~\cite{liu2019unsupervised}. {\it Conceptor negation} is a soft shrinkage of the PCs of a subspace such as stop words or, in our case, of the target words defining the bias directions~\cite{liu2018correcting}. 
Therefore it has the potential to debias better than hard projection (e.g., \citealp{bolukbasi2016man}) while retaining enough semantics. Mathematically, it can capture, conjoin, and negate the bias concepts by logical operation, and thus can \textbf{deal with intersectional bias efficiently}. 

Although \citet{karve2019conceptor} showed that debiasing conceptors can successfully debias both static embeddings such as Glove, Word2vec, and Fasttext, and contextual embeddings such as ELMo~\cite{peters-etal-2018-deep}, they state that the performance in BERT is far less consistent and effective than other word representations. We discover that this is the result of their having selected the wrong set of attribute words, which leads to a poor bias subspace.\footnote{We fixed the coding issues.} Another difference is that the BERT tokens of attribute words should be averaged if they contain multiple subwords after tokenization \citep{liang2020towards,kaneko2021debiasing}.



\section{The Mechanism of Conceptors}\label{subsec:conceptors-math}

\begin{figure}[ht!]
    \centering
    \includegraphics[width=0.18\textwidth]{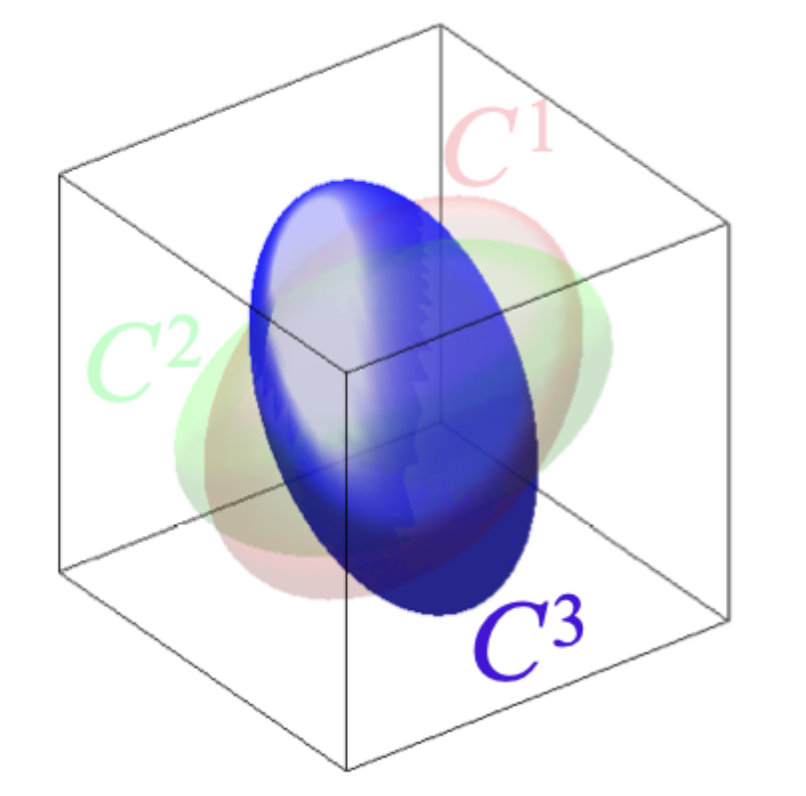}
    \caption{Geometry of three conceptors in the shape of ellipsoids~\cite{jaeger2014controlling}.}
    \label{fig:conceptor-filter}
\end{figure}

Let us take a closer look at the mathematics of conceptors: considering a set of vectors $\{x_{1}, \cdots, x_{n}\}$, $x_i \in \mathbb{R}^N$
for all $i \in \{1, \cdots, n\}$,
a conceptor matrix $C$ is a regularized identity map that minimizes 
\begin{equation}
    \frac{1}{n} \sum_{i=1}^{n}\left\|x_{i}-C x_{i}\right\|_{2}^{2}+\alpha^{-2}\|C\|_{\mathrm{F}}^{2}
    \label{eq:1}
\end{equation}
\noindent
where $\|\cdot\|_\mathrm{F}$ is the Frobenius norm and $\alpha^{-2}$ is a scalar hyper-parameter called \textit{aperture}
\footnote{The default value of $\alpha$ is $1$; we empirically find that grid-searching is not helpful for debiasing so keep it as default}.
It can be shown that $C$ has a closed-form solution:
\begin{equation}
    C=\frac{1}{n} X X^{\top}\left(\frac{1}{n} X X^{\top}+\alpha^{-2} I\right)^{-1} \label{eq:2}
\end{equation}
\noindent
where $X=\left[x_{i}\right]_{i \in\{1, \cdots, n\}}$ is a data collection matrix whose $i$-th column is $x_i$. Intuitively, $C$ is a soft projection matrix on the linear subspace where the typical components of $x_i$ samples lie so that it can capture the components that all representations roughly share. 
Therefore, different from PCA projection which removes the first several principal components (PCs) completely, the conceptors method softly downscales the PCs adjusted by a regularized identity map (Figure~\ref{fig:conceptor-filter}).

\begin{figure}[ht!]
    \centering
    \includegraphics[width=0.45\textwidth]{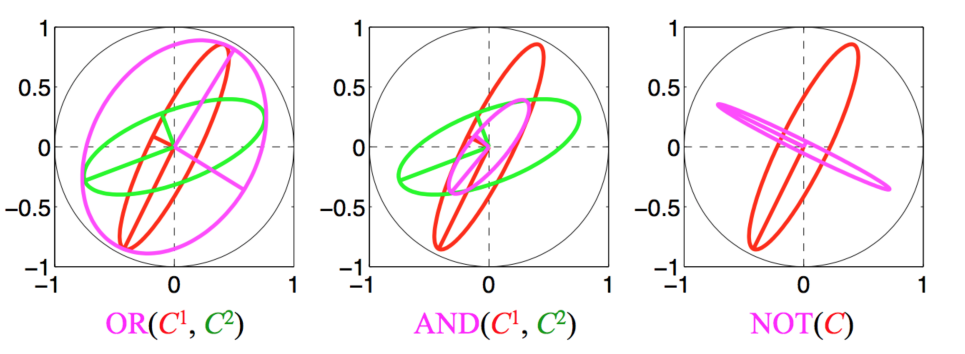}
    \caption{Visualizing the boolean operations on two conceptor matrices. The OR (AND) operator leads to a conceptor matrix (in pink color) with the smallest (largest) ellipsoid~\cite{he2018overcoming}. In our case, it is then negated by the NOT operator to debias.}
    \label{fig:conceptor-logic}
\end{figure}

Conceptors support Boolean operations such as \textbf{NOT} ($\neg$), \textbf{AND} ($\wedge$) and \textbf{OR} ($\vee$). For two arbitrary conceptors $C_1$ and $C_2$, we have
\begin{align}
    \neg C_1 &= I - C_1 \label{eq:3} \\ 
    C_1 \wedge C_2 &= (C_1^{-1} + C_2^{-1} - I)^{-1} \label{eq:4} \\
    C_1 \vee C_2 &= \neg(\neg C_1 \wedge \neg C_2) \label{eq:5} \\ 
    = I - &((I-C_1)^{-1} + (I-C_2)^{-1} - I)^{-1} \nonumber
\end{align}
These logical operations are feasible if $C_1$ and $C_2$ are created by the sets of equal sizes~\cite{jaeger2014controlling}, as shown in Figure~\ref{fig:conceptor-logic}.
This reveals the potential for debiasing by combining different conceptors learned from different bias subspaces. This is helpful both in combining different wordlists for the same bias (e.g. gender) or different wordlists for different protected classes (e.g. gender and race).

\section{Debiasing Sentence Representations}

\subsection{Bias Subspace Setting\label{sec:setting}} 

We explore the impact of different choices of attribute wordlists, the corpora used to find their embeddings, and how the wordlists are combined and filtered to remove outliers, on the quality of bias subspace, and hence the debiasing (Fig~\ref{fig:pipeline}). Different procedures of bias subspace construction yield significantly different debiasing performances.

\paragraph*{Corpora} 
We compare three corpora: 
(1) the \textbf{Brown} Corpus~\cite{francis1979brown}, a collection of text samples of mixed genres; (2) the Stanford Sentiment Treebank (\textbf{SST}; \citealp{socher2013recursive}), a polarized dataset of 10,662 movie reviews; and (3) a \textbf{Reddit} Corpus~\cite{liang2020towards}, a dataset collected from discussion forums about relationships, electronics, and politics.  
The reason is to see how the language formality and topic breadth of texts impact the debiasing, 
the Brown corpus is formal and contains 15 genres, the Reddit corpus is informal with 3 domains and the SST corpus is informal, has only one domain. 
They are used to provide embeddings for the attribute words.

\paragraph*{Combining and Filtering Wordlists} 

We compare five ways of using three different wordlists to create conceptor bias subspaces. 

The three wordlists are gender words originating from different sources: the \textit{pronouns wordlist} is a set of common terms that are specific to particular genders, such as `daughter' or `son'; the  \textit{extended wordlist}, an extension of the former, contains less frequent words such as `cowgirls' or `fiancees'; and \textit{propernouns wordlist} is comprised of proper nouns like `Tawsha', `Emylee', and so on.

There are five methods of using these three wordlists to generate a bias subspace. We can use each of them individually (their subspaces are named the same as themselves: \textbf{pronouns}, \textbf{extended}, and \textbf{propernouns}, respectively).
We can also combine them in two ways: either by concatenating them as a single list generating a corresponding subspace (named \textbf{all}); or by running the conceptor OR operation--a Boolean operation of conceptors described in ~\autoref{subsec:conceptors}--on the three corresponding conceptor matrices to generate what can be viewed as a union of the three bias subspaces (named \textbf{or}). 

\begin{figure}[H]
\centering
    \subfigure[Wordlist]{\centering
    \includegraphics[width=.3\linewidth]{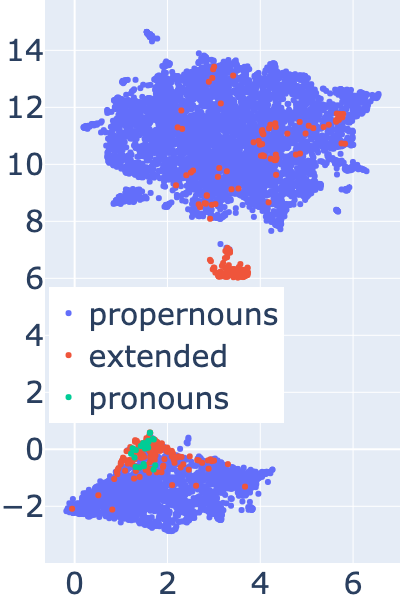}}
    \qquad
    \subfigure[Gender]{\centering
    \includegraphics[width=.3\linewidth]{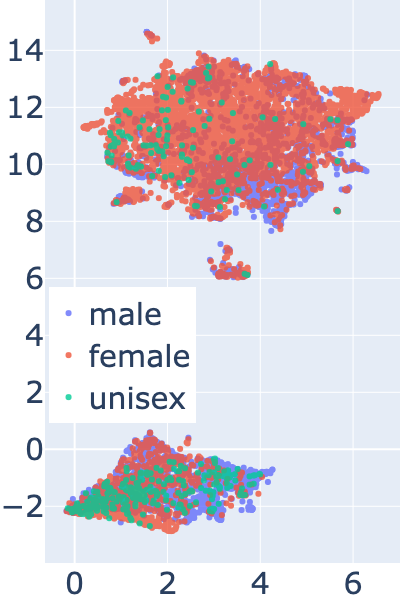}}
\caption{2D UMAP BERT Embeddings of Words.}
\label{fig:umap-wordlist}
\end{figure}

Unlike \citet{karve2019conceptor}, to study the effects of removing outliers from the wordlists, we first project the LLM's embeddings of the words in the wordlist to a 2-dimensional UMAP clustering~\cite{mcinnes2018umap} space, shown in Figure~\ref{fig:umap-wordlist}, and then filter the outliers by percentile on their $(x,y)$-coordinate. The outliers are defined as the points that fall outside of 1.5 times the inter-range (IR), which is the difference between $p$-th and $(1{-}p)$-th percentile.  
We iterate $p$ from $0.1$ to $1.0$ with step size $0.1$ to generate different wordlists and then test how well each debiases. Our goals are to detect the negative effect of outliers on debiasing performance and 
to explore which percentile here is optimal for debiasing.

\subsection{Debiasing Methods}

We propose and explore two kinds of conceptor-aided debiasing: {\it conceptor post-processing}, and {\it conceptor-intervened continued training}. They are abbreviated as P.P. and C.T. respectively in tables. 

\paragraph*{Conceptor Bias Subspace Construction}

We construct the conceptor negation matrix $\neg C$ as demonstrated in Algorithm~\ref{algo},
where matrix $X$ is a stack of the within-sentence contextualized embeddings of the words. 
The words are determined by attribute wordlists and the sentences are from the specified corpus as mentioned in Section~\ref{sec:setting}. 
Note that we do not need the ``difference space'' of bipolar bias as the conceptor projection matrix is applied to the original space--in this way the conceptor method is different from the so-called hard-debiasing~\cite{bolukbasi2016man}.
To ensure contextualization we remove the less-than-four-word sentences. Also, following \citet{kaneko2021debiasing}'s idea, if a word crosses multiple sub-tokens, then its contextualized embedding is computed by averaging the contextualized embeddings of its constituent sub-tokens, which is different than the previous conceptor works. 

\newlength{\commentindent}
\setlength{\commentindent}{.55\textwidth} 
\makeatletter
\renewcommand{\algorithmiccomment}[1]{\unskip\hfill\makebox[\commentindent][l]{//~#1}\par}
\LetLtxMacro{\oldalgorithmic}{\algorithmic}
\renewcommand{\algorithmic}[1][0]{%
  \oldalgorithmic[#1]%
  \renewcommand{\ALC@com}[1]{%
    \ifnum\pdfstrcmp{##1}{default}=0\else\algorithmiccomment{##1}\fi}%
}
\makeatother

\begin{figure*}[ht]
    \begin{minipage}{\linewidth}
    \begin{algorithm}[H]
    \small
    \caption{\label{algo} \textsc{Conceptor-Debias}: a conceptor-aided post-process algorithm for debiasing LLMs.}
    \begin{algorithmic}[1]
        \REQUIRE large language model $\mathbbmsl{M_\theta}$ (with parameters $\theta$), bias attribute wordlist $\mathcal{W}$, and corpus $\mathcal{S}$. 
        \STATE $X$ $\gets$ [ ]
        \FOR{each word $w \in \mathcal{W}$}
            \FOR{each sentence $s \in \mathcal{S}$}
                \IF{$w$ inside $s$}
                    \STATE $w_{c} \gets$ the embedding of $w$ inside $\mathbbmsl{M_\theta}(s)$\ \ \ \ \COMMENT{get contextualized word embedding}
                    \STATE $X \gets X + w_{c}$ \COMMENT{stack as a matrix} 
                \ENDIF
            \ENDFOR
        \ENDFOR
        \STATE $C \gets X X^{\top}( X X^{\top} + I)^{-1}$ \COMMENT{construct conceptor bias subspace} 
            \COMMENT{note that different $X_i$ yields different $C_i$ for arbitrary $i$}
        \STATE $C \gets (C_1^{-1} + C_2^{-1} - I)^{-1}$ \COMMENT{cross bias subspaces by AND operator (if intersectional debias)} 
        \STATE $C \gets I - ((I-C_1)^{-1} + (I-C_2)^{-1} - I)^{-1}$ \COMMENT{unite bias subspaces by OR operator (for robust debias)}
        \STATE $\neg C \gets I - C$ \COMMENT{make negation conceptor matrix by NOT operator}
        \FOR{each new sentence $t$}
            \STATE $t^* \gets \neg C \cdot \mathbbmsl{M_\theta}(t)$ \COMMENT{debias sentence by projection}
        \ENDFOR
    \end{algorithmic}
    \end{algorithm}
    \end{minipage}
\end{figure*}

\paragraph*{Conceptor Negation and Post-Processing} 
Next, we post-process the sentence embeddings $t$ which contain attribute words and target words, by taking the matrix product of $\neg C$ to subtract the bias subspace, rendering debiased embeddings $t^*$, as demonstrated in the last part of Algorithm~\ref{algo}.
Each BERT layer manifests different levels of bias \citep{bommasani2020interpreting}. To maximize the effectiveness of $\neg C$, we want $\neg C$ to be generated from the corresponding  layer. Therefore, we are the first ones to test the debiasing performance by using different conceptor matrices generated by different layers of the language model and to explore whether conceptor post-processing generalizes well on each layer of LLMs and on different LLMs (BERT and GPT). 

\paragraph*{Intersectioanl Debiasing} Importantly, not only can conceptors mitigate different types of biases such as gender and race respectively, but it can also conjoin and negate these biased concepts together due to its magical logical operations. It is natural that societal biases co-exist in multi-dimensions: such as ``African Male'' rather than ``African'' and ``Male''. Therefore, it is efficient that conceptors can tackle them intersectionally by utilizing the previously constructed bias subspaces via its OR operation to construct the new mixed conceptors.

\begin{figure}[h]
    \centering
    \includegraphics[scale=0.45]{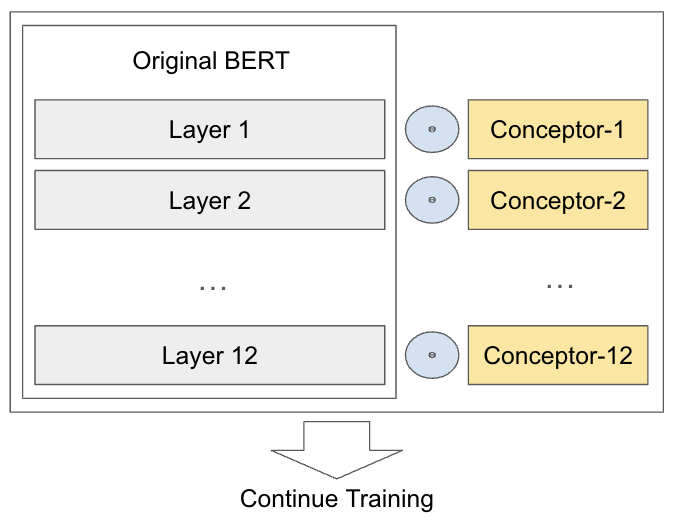}
    \caption{Conceptor-Intervened BERT (CI-BERT). Each model's layer takes the matrix product (blue circle) of the conceptor-X generated from the corresponding layer X. It can be used directly or continually trained. 
    }
    \label{fig:arch}
\end{figure}

\paragraph*{Conceptor Intervention and Continued Training}  
The varying levels of bias across BERT layers suggest the possible utility of an alternate approach to mitigate the bias. 
Accordingly, we construct a new architecture, \textit{Conceptor-Intervened BERT} (CI-BERT), by placing the corresponding conceptor matrix after each layer of BERT (Figure~\ref{fig:arch}). We then continue training the entire model to incorporate the model weights with the bias negation captured by the conceptors in each layer. Thus we can take the biases in all layers into account so that we can mitigate the layerwise bias simultaneously. 

CI-BERT architecture can be used in three ways. 
We can load the original pre-trained weights to CI-BERT and directly render the language embeddings (\textbf{Type I}; CI-BERT $\times$ original weights). Alternatively, we can continue training the model using CI-BERT to get newly trained weights; then we can load these weights back to either the original off-the-shelf BERT architecture (\textbf{Type II}; BERT $\times$ trained weights) or to the new architecture CI-BERT (\textbf{Type III}; CI-BERT $\times$ trained weights). 





\section{Quantifying Bias\label{sec:evaluate-bias}}

\subsection{Sentence Encoder Association Test} 
The Sentence Encoder Association Test (SEAT) \citep{may2019seat} is an extension of the Word Encoder Association Test (WEAT)~\cite{caliskan2017semantics}. It can measure the bias at the sentence level in different kinds of bias~\cite{meade_2022_empirical}.

SEAT uses  two types of words: \textit{attribute} words $\mathcal{W}_a$ (e.g. he/she) and \textit{target} words $\mathcal{W}_t$ (e.g. occupations), which we expect to be gender-neutral. 
That is, the associations between $w_a/w_a^\prime \in \mathcal{W}_a$ and $w_t \in \mathcal{W}_t$ 
should be no difference in the sentence-template representations of LLMs.

Denote the sentence sets of attribute words as $A$ and $A^\prime$, and of target words as $T$ and $T^\prime$, we have:
\begin{equation*} \scalemath{0.9}{
c(A, A^\prime, T, T^\prime) = \sum_{t \in T} c(t, A, A^\prime) - \sum_{t^\prime \in T^\prime} c(t^\prime, A, A^\prime) 
} \end{equation*}
\noindent
where for each sentence $s$, we have $c(s, A, A^\prime)$, the difference of the mean cosine similarity of $s$ concerning sentences from between $A$ and $A^\prime$; as
\begin{equation*} \scalemath{0.9}{
c(s, A, A^\prime)=\frac{1}{|A|} \sum_{a \in A} \cos (s, a) - \frac{1}{|A^\prime|} \sum_{a^\prime \in A^\prime} \cos (s, a^\prime)
} \end{equation*}
\noindent
The amount of bias is given by the effect size
\begin{equation*} \scalemath{0.9}{
d=\frac{\mu\left(\{c(t, A, A^\prime)\}_{t \in T}\right)-\mu\left(\{c(t^\prime, A, A^\prime)\}_{t^\prime \in T^\prime}\right)}{\sigma\left(\{c(a, T, T^\prime)\}_{a \in A \cup A^\prime}\right)}
} \end{equation*} 
\noindent
where $\mu$ and $\sigma$ denote the mean and standard deviation, respectively. The smaller the absolute value of $d$ is, the less bias has been detected. 
The one-sided p-value measures the likelihood that a random resampling of the sentence set that contains attribute words would generate the observed test statistic.

\subsection{Gender Co-Reference Resolution} 
As described by~\citet{gonen-goldberg-2019-lipstick}, SEAT can detect only the presence but not the absence of bias. To further understand how the conceptor-aided methods work on debiasing, we adopt an end-task: gender co-reference resolution. 

WinoBias~\citep{zhao2018gender} provides gender-balanced co-reference tests to evaluate LLMs' neutrality towards pronouns referring to occupations. The tests include pro-stereotypical (PRO) scenarios, where gender pronouns match gender-conforming occupations (e.g., her/nurse), and anti-stereotypical (ANTI) scenarios, where gender pronouns apply to disfavored occupations. The bias is measured by the average and absolute difference in F1 scores between the PRO and ANTI subsets.

Based on this, \citet{Manela2021StereotypeAS} develop two intuitive metrics, \textit{skew} and \textit{stereotype}, to better probe model fairness. In formula,
\begin{equation*} \scalemath{0.9}{
    \begin{aligned}
    \mu_{\text {Skew }} & \triangleq \frac{\left|\mathrm{~F} 1_{\text {pro }}^{M}-\mathrm{F} 1_{\text {pro }}^{F}\right| + 
    \left|\mathrm{F} 1_{\text {anti }}^{M}-\mathrm{F} 1_{\text {anti }}^{F}\right|}{2} \\
    \mu_{\text {Stereo }} & \triangleq \frac{\left|\mathrm{~F} 1_{\text {pro }}^{M}-\mathrm{F} 1_{\text {anti }}^{M}\right| + 
    \left|\mathrm{F} 1_{\text {pro }}^{F}-\mathrm{F} 1_{\text {anti }}^{F}\right|}{2}
    \end{aligned}
} \end{equation*}
\noindent
where superscripts $M$ and $F$ denote male and female respectively and $\mathrm{~F}1$ stands for the $\mathrm{~F}1$-score. It is shown that there is an approximate trade-off between these two biases. The authors argue that the T2 test set of WinoBias is better than the T1 test set at revealing bias, as the latter is less ambiguous to LLMs. Therefore, we only report T2 here.

\section{Debiasing Results}

This section aims to answer these questions:
\begin{itemize}[noitemsep,topsep=0pt]
\item What is the best setting for bias subspace generation within conceptor-aided debiasing? 
\item Given the best setting, can the conceptor post-processing mitigate bias and beat SoTA? 
\item Does embedding conceptors into LLMs via continued training beat post-processing? 
\item What roles can conceptors operators--NOT, OR, AND--play in the debiasing pipeline?
\end{itemize}

To help comparison, the SoTA debiasing results from~\citet{meade_2022_empirical} is included in the tables. 

\subsection{Models}
To investigate the generalization of conceptor debiasing, we explored different scales of typical LLM families, which are: BERT-T (\textit{bert-tiny}), BERT (\textit{bert-base-uncased}), BERT-L (\textit{bert-large-uncased}), GPT2 (\textit{gpt2}), GPT-L (\textit{gpt2-large}), and GPT-J (\textit{gpt-j}). 
We did not test on GPT3 and ChatGPT since their embedding models (e.g. \textit{text-embedding-ada-002}) do not support the contextualized embedding on token level. 
However, due to the similar modeling, once we have such embedding, conceptor techniques can be transferred.

\subsection{Bias Subspace Construction with Robustness Boosted via OR Operator}

We construct the conceptor bias subspaces upon the different combinations of corpora, wordlist selections, and outlier removal.

To evaluate corpora, by testing on the last layer of the BERT, we compare the debiasing result of three different corpora: Brown, SST, and Reddit on SEAT. Table~\ref{tab:seat-belt-different-corpus} shows that Brown stably provides the best debiasing result even if using different wordlist subspaces. The SST corpus is a close second, while Reddit is by far the worst. The style of the Reddit corpus is most likely least similar to that of the SEAT evaluations.

To evaluate alternate methods of constructing the bias wordlist subspace, we use the five subspaces described in Section~\ref{sec:setting}. Among them, the \textit{or} subspace is the most robust; see Table~\ref{tab:seat-belt-different-percentiles-on-brown}, \ref{tab:seat-belt-different-percentiles-on-sst} and \ref{tab:seat-belt-different-percentiles-on-reddit}. 
Combining the \textit{pronouns}, \textit{extended} and \textit{propernouns} subspaces with \textit{or} represents the distinct yet union concepts (and hence subspaces) of each of the wordlists, thus both outperforming individual wordlists and outperforming the \textit{all} subspace, which simply concatenates all the wordlists, giving a less precise subspace definition.

To evaluate wordlist outlier removal, we define the outliers by the UMAP filter as discussed in section~\ref{sec:setting} and generate different percentages of the words that are used to capture bias. For example, the \textit{all} subspace has $2071$  words within $0.5{-}1.0$ percentile, $2061$ in the $0.4$ percentile, $1601$ in the $0.3$ percentile, $430$ in the $0.2$ percentile, and $82$ in the $0.1$ percentile (Table~\ref{tab:num-word-percentile}). We observe that including fewer words often leads to higher debiasing performance, presumably due to the removal of outliers. However, an extremely small percentile, say $0.1$, would harm the effectiveness of debiasing because of the inadequate loss being left (Table~\ref{tab:seat-belt-different-percentiles-on-brown}, \ref{tab:seat-belt-different-percentiles-on-sst} and \ref{tab:seat-belt-different-percentiles-on-sst}). Similar results are obtained if using T-SNE~\cite{van2008visualizing}.

In conclusion, the optimal setting for BERT-T is ``sst-0.5-or'' (SST; percentile 0.5; \textit{or} subspace); 
similarly, for BERT is ``brown-0.4-or'' (Brown; percentile 0.4; \textit{or} subspace). 
For other models, if not mentioned, it is \textit{default} as ``brown-1.0-or''.
Henceforth, these settings are held for the conceptor debiasing on the models respectively.

\subsection{Post-Processing Debias via NOT Operator}
For general debiasing via conceptor negation post-processing, the performance is excellent. The SEAT score of BERT decreases from $0.620$ to around $0.350{-}0.400$ in Brown Corpus (Table~\ref{tab:seat-belt-different-percentiles-on-brown}), and can be as low as $0.311$ if using the setting ``brown-0.4-or'', outperforming the debiasing result of CDA, \textsc{Dropout} and \textsc{\textsc{SentenceDebiase}} (Table~\ref{tab:seat-bert-and-llms}). 
The success of debiasing is further verified by WinoBias (Table~\ref{tab:winobias-skew-stereo}), where the skew bias drops from $38.3$ to $22.3$ without any additional fine-tuning. Although the stereotype bias increases, it is not only expected since these two biases are trade-offs but also acceptable, as they now reach a good balance~\cite{Manela2021StereotypeAS}.

The debiasing conceptors are robust and generalizable, as shown in Table~\ref{tab:seat-bert-and-llms}, the debias performance is consistent in different scales of BERT and GPT models. Note that the settings of BERT-L, GPT2-L and GPT-J are not \textit{tuned} (i.e. \textit{default} setting), which means that they can likely reach much lower SEAT scores.
Moreover, conceptors can mitigate the bias in almost all scenarios, 
no matter using which corpus, bias subspace, or wordlist threshold (Table~\ref{tab:seat-belt-different-percentiles-on-brown}, 
\ref{tab:seat-belt-different-percentiles-on-reddit} and 
\ref{tab:seat-belt-different-percentiles-on-sst}); 
no matter which LLMs 
(Table~\ref{tab:seat-bert-and-llms}, 
~\ref{tab:seat-bert-tiny-different-percentiles-on-brown}, 
\ref{tab:seat-bert-tiny-different-percentiles-on-reddit}, 
\ref{tab:seat-bert-tiny-different-percentiles-on-sst} and 
,\ref{tab:gpt2-different-percentiles-on-brown})
;
no matter in which layer 
(Table~\ref{tab:seat-belt-different-layer-conceptor-intervened-brown}, 
\ref{tab:seat-belt-different-layer-conceptor-intervened-sst} and 
\ref{tab:seat-bert-tiny-different-layer-conceptor-intervened});
and no matter which type of biases
(Table~\ref{tab:intersect}, Table~\ref{tab:intersect-gender} and~\ref{tab:intersect-race}).

\begin{table*}[ht!]
\centering
\small
\begin{tabular}{lllllllr}
\hline Model & SEAT-6 & SEAT-6b & SEAT-7 & SEAT-7b & SEAT-8 & SEAT-8b & Gender (AAvg.) \\
\hline 
BERT & 0.931$^*$ & 0.090 & -0.124 & 0.937$^*$ & 0.783$^*$ & 0.858$^*$ & 0.620 \\
+ Conceptor P.P. (tuned) & \textbf{0.388} & \textbf{-0.078} & -0.292 & \textbf{0.179} & \textbf{0.594}$^*$ & \textbf{0.335} & \ColorDown{0.309} \textbf{0.311} \\
+ Conceptor C.T. & \textbf{0.227} & 0.426 & -0.341 & \textbf{-0.253} & \textbf{-0.344} & \textbf{-0.088} & \ColorDown{0.340} \textbf{0.280} \\
\hdashline
+ CDA & \textbf{0.846}$^*$ & 0.186 & -0.278 & 1.342$^*$ & 0.831$^*$ & \textbf{0.849}$^*$ & \ColorUp{0.120} 0.722 \\
+ \textsc{Dropout} & \textbf{1.136}$^*$ & 0.317 & 0.138 & 1.179$^*$ & 0.879$^*$ & 0.939$^*$ & \ColorUp{0.144}  0.765 \\
+ INLP & \textbf{0.317} & -0.354 & -0.258 & \textbf{0.105} & \textbf{0.187} & \textbf{-0.004} & \ColorDown{0.416} \textbf{0.204} \\
+ \textsc{SentenceDebias} & \textbf{0.350} & -0.298 & -0.626 & \textbf{0.458}$^*$ & \textbf{0.413} & \textbf{0.462}$^*$ & \ColorDown{0.186} \textbf{0.434}\\
\hline
\hline
BERT-L & 0.370 & -0.015 & 0.418$^*$ & 0.221 & -0.258 & 0.711$^*$ & 0.332 \\
+ Conceptor P.P. (default) & \textbf{0.197} & -0.206 & 0.064 & \textbf{0.065} & -0.371 & \textbf{0.337} & \ColorDown{0.125} \textbf{0.207}  \\
\hline
GPT2 & -0.510 & 0.057 & -0.274 & -0.186 & -0.369 & -0.313 & 0.285 \\
+ Conceptor P.P. (tuned) & \textbf{0.092} & 0.316 & \textbf{-0.001} & \textbf{0.064} & \textbf{-0.035} & \textbf{-0.062} & \ColorDown{0.190} \textbf{0.095} \\
\hline
GPT2-L & 1.093$^*$ & 0.192 & 0.214 & 1.354$^*$ & 0.861$^*$ & 1.157$^*$ & 0.812 \\
+ Conceptor P.P. (default) & \textbf{1.055}$^*$ & \textbf{0.008} & \textbf{-0.089} & 1.406$^*$ & \textbf{0.282} & \textbf{0.992}$^*$ & \ColorDown{0.173} \textbf{0.639} \\
\hline
GPT-J & 1.299$^*$ & 0.300 & 0.962$^*$ & 1.434$^*$ & 0.617$^*$ & 1.031$^*$ & 0.940 \\
+ Conceptor P.P. (default) & \textbf{1.184}$^*$ & \textbf{0.285} & \textbf{0.661}$^*$ & \textbf{1.284}$^*$ & \textbf{0.558}$^*$ & \textbf{1.024}$^*$ & \ColorDown{0.107} \textbf{0.833} \\
\hline
\end{tabular}
\caption{\label{tab:seat-bert-and-llms} SEAT effect size of gender debiased BERT and GPT model. Effect sizes closer to 0 indicate less biased sentence representations (\textbf{bolded value}). Statistically significant effect sizes at $p<0.01$ are denoted by *. The final column is the average absolute SEAT score of the first six columns. 
\textit{Default} means using the default setting: brown corpus, no wordlist filtering, and OR subspace; while \textit{tuned} means using the optimal combination of corpus, wordlist percentile, and conceptor bias subspace.
P.P. stands for post-processing, while C.T. stands for continued training. 
The full version is in Appendixes~\ref{app:additional-bert-base} and \ref{app:additional-gpt2}.\\ 
}
\end{table*}

\begin{table}[H]
    \centering
    \scriptsize
    \begin{tabular}{lcccc:cc}
    \hline \multirow{2}*{Model}  & \multicolumn{2}{c}{ F1 Male } & \multicolumn{2}{c}{ F1 Female} & \multicolumn{2}{c}{ Bias } \\
    \cline { 2 - 7 } & Pro & Anti & Pro & Anti & Stereo & Skew \\
    \hline BERT       & 66.4 & 58.9 & 31.8 & 17.0 & 11.2 & 38.3 \\
     + Conceptor P.P. & 69.5 & 48.1 & 52.8 & 20.1 & 27.0 & \textbf{22.3} \\
     + Conceptor C.T. & 41.0 & 39.3 & 57.6 & 56.6 & \textbf{4.1} & \textbf{17.0} \\
    \hline
\end{tabular}
\caption{\label{tab:winobias-skew-stereo} F1 of skew and stereotype biases in WinoBias.}
\end{table}

\subsection{Intersectional Debias via AND Operator}
Table~\ref{tab:intersect} empirically shows that conceptors not only can mitigate the different type of biases, but also can intersect the existing bias subspaces (e.g. gender and race) to create a mixed conceptor matrix in an efficient way and to debias gender and race respectively.
Furthermore, for assessing the intersectional debiasing, we employ the I1-I5 intersectional bias test introduced by \citet{tan2019assessing}. They adapt the SEAT to examine the privilege associated with the combination of being African/European American and being male or female. 
The results demonstrate that such intersected conceptor formed via the AND operator can effectively reduce multi-dimensional bias, lowering the SEAT score from $0.673$ to $0.434$, while its conceptor counterparts focused solely on single-dimensional bias can only reduce the score to $0.613$ and $0.635$ respectively.

\begin{table*}[ht!]
    \centering
    \footnotesize
    \setlength{\tabcolsep}{2.4pt}
    \begin{tabular}{l:rr:cccccr}
    \hline 
    Model & Gender (AAvg.) & Race (AAvg.)
        & SEAT-I1 & SEAT-I2 & SEAT-I3 & SEAT-I4 & SEAT-I5 & Intersect (AAvg.)  \\
    \hline 
    BERT & 0.620 & 0.620 
        & 0.389$^*$ & -0.424 & 1.195$^*$ & 0.525$^*$ & 0.834$^*$ & 0.673 \\
    + Gender Conceptor & \ColorDown{0.309} \textbf{0.311} & N/A 
        & 0.394$^*$ & -0.456 & \textbf{1.156}$^*$ & \textbf{0.413}$^*$ & \textbf{0.755}$^*$ & \ColorDown{0.060} \textbf{0.613} \\
    + Race Conceptor & N/A & \ColorDown{0.043} \textbf{0.577} 
        & 0.394$^*$ & -0.456 & \textbf{1.156}$^*$ & \textbf{0.413}$^*$ & \textbf{0.755}$^*$ & \ColorDown{0.062} \textbf{0.635} \\
    + Intersect Conceptor$^\mathsection$ & \ColorDown{0.029} \textbf{0.591} & \ColorDown{0.016} \textbf{0.604} 
        & \textbf{0.214} & -0.474 & \textbf{0.872}$^*$ & \textbf{0.207} & \textbf{0.403}$^*$ & \ColorDown{0.239} \ColorBest{\textbf{0.434}} \\
    \hline
\end{tabular}
\caption{\label{tab:intersect} SEAT effect size of race, gender, and intersectionally debiased BERT model, where the absolute average SEAT score of gender, race, and intersect are across 6, 7, 5 tests, respectively. The full version is in Appendix~\ref{app:intersect}.\\
\small{$^{\mathsection}$ It indicates the conceptor matrix generated by its negated AND operation of gender conceptor matrix and race conceptor matrix}}
\end{table*}

\subsection{Conceptor-Intervention Debias}

We use CI-BERT architecture to continue to train the models to get the new weights. Then we demonstrate the combinations of architectures and weights as an ablation study (Type I, II, and III). Among them, Type III can outperform conceptor post-processing (Table~\ref{tab:seat-bert-and-llms}), and Type I and II (Table~\ref{tab:ci-bert-ablation}). 

Compared to the SEAT score after post-processing, Type I can outperform it at each layer of BERT-T but underperform it at most layers of BERT (Table~\ref{tab:seat-belt-different-layer-conceptor-intervened-sst} and \ref{tab:seat-bert-tiny-different-layer-conceptor-intervened}). 

\begin{table}[h]
    \centering
    \footnotesize
    \begin{tabular}{c|cc|c}
    \hline Type & CI-BERT (Arch.) & Trained Weights & SEAT  \\
    \hline (Orig.) &            &            & 0.620 \\
           I       & \checkmark &            & 0.336 \\
           II      &            & \checkmark & 0.592 \\
           III     & \checkmark & \checkmark & 0.280 \\
    \hline
\end{tabular}
\caption{\label{tab:ci-bert-ablation} The ablation study of architecture and weights of CI-BERT evaluated by SEAT (the same as Table~\ref{tab:seat-bert-and-llms}).}
\end{table}

In short, using the CI-BERT with the newly trained weights could receive the lowest bias in the model and is promising to beat post-processing. 
For example, when using the setting ``brown-0.4-or'', the lowest SEAT score is $0.280$, beating the post-processing result of $0.311$ and more than half of the SoTA methods. This is verified again by gender co-reference resolution in Table~\ref{tab:winobias-skew-stereo}--in comparison to its post-processing counterpart, CI-BERT continued training lowers both stereotype bias by $22.9$ and skew bias by $5.3$ from Test Set 2 of WinoBias. This is non-trivial since these two biases are a tradeoff and thus generally hard to decrease simultaneously~\cite{Manela2021StereotypeAS}.

\begin{figure}[ht]
\centering
\begin{subfigure}{
  \centering
  \includegraphics[width=.8\linewidth]{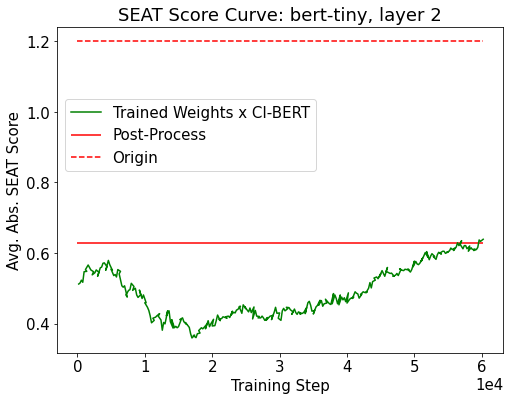}}
\end{subfigure}
\begin{subfigure}{
  \centering
  \includegraphics[width=.8\linewidth]{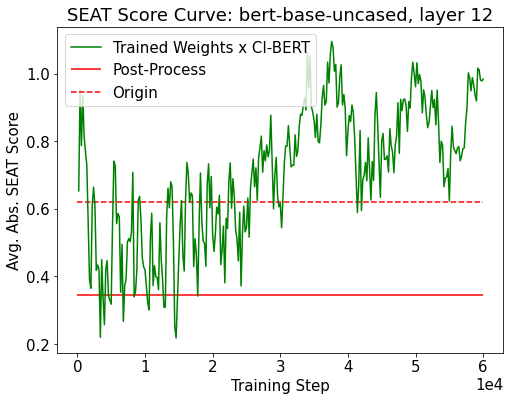}}
\end{subfigure}
\caption{SEAT score curve of CI-BERT continued training. We compare the results with the original embeddings and post-processed embeddings. We test on the last layer of BERT-T (top) and BERT (bottom).}
\label{fig:seat-score-curve}
\vspace{-20pt}
\end{figure}

To further study the feasibility and robustness of CI-BERT continued training concerning the model property. We experiment on both BERT-T and BERT and plot the average SEAT curve along with training steps (Figure~\ref{fig:seat-score-curve}). Both can beat their post-processing counterparts in some steps during the early training stage, and then the bias fluctuates and increases again, perhaps due to the model relearning the bias during continued training, or oversaturating the conceptor bias projections into its weights. 

In comparison, the continual-trained CI-BERT can more stably lower the bias in smaller Bert model. We suspect this is related to the model complexity. 
The debiasing projection of the last layer’s conceptor matrix is upon the last hidden state and thus generated transitively from all the prior layers. Currently, we are embedding all layers’ conceptor matrices, which may lead to overlapping and redundant debiasing projection from the prior layers. 

\subsection{Maintaining Meaningful Semantics\label{sec:maintain-semantics}}

To understand how conceptor debiasing impacts the downstream natural language understanding (NLU) tasks, the GLUE benchmark~\citep{wang2018glue}--comprised of nine different tasks--is used to evaluate the model after debiasing (Table~\ref{tab:glue}). 
While there seems to be no consensus about the quantitative threshold of the trade-off between language modeling capability and debiasing performance, a small decrease may be acceptable depending on the downstream tasks. We believe that, in an ideal scenario, the performance on the GLUE benchmark should not significantly decline after debiasing.

The conceptor post-processing of BERT can retain and even improve the useful semantics (increase the average GLUE score by 1.77) for downstream tasks without damage to the model's ability, outperforming any other listed SoTA debiasing methods. Even if scaling to BERT-L, the GLUE is still slightly higher. In comparison, the average GLUE score of conceptor continued-training BERT is relatively low, although it is not the worst among all the methods. This indicates that the continued-training method, while still capable of outperforming its post-processing counterpart under the same  setting, may sacrifice NLU abilities. 

Since GPT is an autoregressive model, we adopt the SequenceClassification counterpart on the GLUE benchmark, following the method of \citet{meade_2022_empirical}. The score of GPT2 and GPT-J are decreased slightly by 0.11-0.14, which is an affordable cost, while GPT2-L increases slightly by 0.05.

\begin{table}[H]
    \centering
    \small
    \begin{tabular}{lr}
\hline Model & Average \\
\hline BERT                          & 77.74 \\
+ Conceptor P.P.                     & \ColorUpCR{1.77} 79.51 \\
+ Conceptor C.T.                     & \ColorDownCR{1.03} 76.71 \\
\hdashline
+ CDA                                & \ColorDownCR{0.22} 77.52 \\
+ \textsc{Dropout}                   & \ColorDownCR{1.46} 76.28 \\
+ INLP                               & \ColorDownCR{0.99} 76.76 \\
+ \textsc{SentenceDebias}            & \ColorUpCR{0.07} 77.81 \\
\hline
BERT-L                               & 78.81 \\
+ Conceptor P.P.                     & \ColorUpCR{0.05} 78.86 \\
\hline
GPT2                                 & 73.01 \\
+ Conceptor P.P.                     & \ColorDownCR{0.11} 72.90 \\
\hline
GPT2-L                               & 75.84 \\
+ Conceptor P.P.                     & \ColorUpCR{0.05} 75.89 \\
\hline
GPT-J                                 & 78.22 \\
+ Conceptor P.P.                     & \ColorDownCR{0.14} 78.06 \\
\hline
\end{tabular}
    \caption{GLUE validation set results for gender debiased BERT and GPT model. The full version is in Appendixes~\ref{app:additional-bert-base} and \ref{app:additional-llms}.} 
    \label{tab:glue}
    \vspace{-10pt}

\end{table}

Notice that even when trained on the original BERT architecture, the average GLUE score still drops about $0.3$ point. Thus, the lower GLUE score here is not completely caused by CI-BERT, though the actual reason is hard to determine due to training randomness~\cite{mccoy2019berts}.

\section{Conclusion and Future Work}

We have shown that conceptor-aided debiasing can successfully mitigate bias from LLMs (e.g., BERT, GPT) by its NOT operation. Specifically,  conceptor post-processing outperforms many state-of-the-art debiasing methods in both debiasing effectiveness and semantic retention. We also tested a new architecture, conceptor-intervened BERT (CI-BERT), which in combination with continued training, takes all layers' biases into account and shows the promise to outperform its post-processing counterpart. However, it might be at the cost of increased instability and worse semantic retention. In all cases, the best conceptor matrices are generally obtained when the bias subspace is constructed using (1) a cleaner corpus, (2) the union of different related wordlists (e.g. pronouns, roles, and names) by the conceptor OR operation, and (3) removal of outliers from the wordlists. 
We further show that cocneptor-aided debiasing is robust in different LLMs, various layers of models, and varied types of biases. Moreover, conceptors can utilize the current conceptor matrices to construct a new conceptor matrix to mitigate the intersectional bias in an efficient manner by AND operation.

In future research we plan to make CI-BERT and intersectional conceptors more robust and effective. 

\section*{Limitations} 

We list several  limitations of our work below.

\paragraph*{1) We only test the binary bias.} We only test the bias in pairs via SEAT and WinoBias, for example, `male'/`female' or `young'/`old'. However, it is widely recognized that terms in gender, race, etc. can be multi-polar. 

\paragraph*{2) Our result is limited to English, and both corpora and wordlist tend towards North American social biases.} The whole of our experiment is conducted in English. In addition, Brown and SST Corpora are collected entirely in the North American environment. So are the wordlists. Therefore, it is expected that they skew towards North American social biases. When such models are debiased under the North American environment, it is necessary to understand how effective they are when transferred to other cultures.

\paragraph*{3) The generalization of conceptor-aided debiasing techniques can be tested more exhaustively.} This work has tested it on gender and race, but it can also be tested on other types of bias such as religious bias and hate speech.

\section*{Ethical Considerations}

 The definition and recognition of bias are subtle. For example, we have used simple traditional binary definitions of male and female to examine gender bias. This, of course, ignores a much wider variety of gender identities, thus introducing an implicit bias to the analysis. Similarly, studies on racial bias rely on possibly problematic definitions of race.  Core to our debiasing method is the selection of the wordlists. Each wordlist carries its own implicit definitions of gender, race, and other important dimensions. Care should be used to ensure that they represent the desired categories.  To this end, it is often useful to involve people from the communities whose language is being debiased to better represent their values and belief systems.
 
 One should also be careful in the use of debiasing.  Removing signals about race or gender is often beneficial in reducing discrimination or in producing better translations. It may also remove key features of models needed for analyses. For example, removing gender or race `signal' from the model may severely hamper the use of that model in gender studies or work on critical race theory.  ``White-washing'' models are not always a benefit; sometimes one wants to see the bias inherent in a corpus.
 
\section*{Acknowledgements}

First, we would like to thank the reviewers for their fruitful discussion with us. We would also like to thank Claire Daniele for her editorial support. Last but not least, we appreciate PennNLP members for their helpful comments.

\bibliography{anthology}

\begin{thebibliography}{42}
\expandafter\ifx\csname natexlab\endcsname\relax\def\natexlab#1{#1}\fi

\bibitem[{Antoniak and Mimno(2021)}]{mimno2021bad}
Maria Antoniak and David Mimno. 2021.
\newblock \href {https://doi.org/10.18653/v1/2021.acl-long.148} {Bad seeds:
  Evaluating lexical methods for bias measurement}.
\newblock In \emph{Proceedings of the 59th Annual Meeting of the Association
  for Computational Linguistics and the 11th International Joint Conference on
  Natural Language Processing (Volume 1: Long Papers)}, pages 1889--1904,
  Online. Association for Computational Linguistics.

\bibitem[{Barikeri et~al.(2021)Barikeri, Lauscher, Vuli{\'c}, and
  Glava{\v{s}}}]{barikeri-etal-2021-redditbias}
Soumya Barikeri, Anne Lauscher, Ivan Vuli{\'c}, and Goran Glava{\v{s}}. 2021.
\newblock \href {https://doi.org/10.18653/v1/2021.acl-long.151}
  {{R}eddit{B}ias: A real-world resource for bias evaluation and debiasing of
  conversational language models}.
\newblock In \emph{Proceedings of the 59th Annual Meeting of the Association
  for Computational Linguistics and the 11th International Joint Conference on
  Natural Language Processing (Volume 1: Long Papers)}, pages 1941--1955,
  Online. Association for Computational Linguistics.

\bibitem[{Bolukbasi et~al.(2016)Bolukbasi, Chang, Zou, Saligrama, and
  Kalai}]{bolukbasi2016man}
Tolga Bolukbasi, Kai-Wei Chang, James~Y Zou, Venkatesh Saligrama, and Adam~T
  Kalai. 2016.
\newblock Man is to computer programmer as woman is to homemaker? debiasing
  word embeddings.
\newblock \emph{Advances in neural information processing systems}, 29.

\bibitem[{Bommasani et~al.(2020)Bommasani, Davis, and
  Cardie}]{bommasani2020interpreting}
Rishi Bommasani, Kelly Davis, and Claire Cardie. 2020.
\newblock Interpreting pretrained contextualized representations via reductions
  to static embeddings.
\newblock In \emph{Proceedings of the 58th Annual Meeting of the Association
  for Computational Linguistics}, pages 4758--4781.

\bibitem[{Brown et~al.(2020)Brown, Mann, Ryder, Subbiah, Kaplan, Dhariwal,
  Neelakantan, Shyam, Sastry, Askell et~al.}]{brown2020language}
Tom Brown, Benjamin Mann, Nick Ryder, Melanie Subbiah, Jared~D Kaplan, Prafulla
  Dhariwal, Arvind Neelakantan, Pranav Shyam, Girish Sastry, Amanda Askell,
  et~al. 2020.
\newblock Language models are few-shot learners.
\newblock \emph{Advances in neural information processing systems},
  33:1877--1901.

\bibitem[{Caliskan et~al.(2022)Caliskan, Ajay, Charlesworth, Wolfe, and
  Banaji}]{Caliskan2022GenderBI}
Aylin Caliskan, Pimparkar~Parth Ajay, Tessa E.~S. Charlesworth, Robert Wolfe,
  and Mahzarin~R. Banaji. 2022.
\newblock Gender bias in word embeddings: A comprehensive analysis of
  frequency, syntax, and semantics.
\newblock \emph{ArXiv}, abs/2206.03390.

\bibitem[{Caliskan et~al.(2017)Caliskan, Bryson, and
  Narayanan}]{caliskan2017semantics}
Aylin Caliskan, Joanna~J Bryson, and Arvind Narayanan. 2017.
\newblock Semantics derived automatically from language corpora contain
  human-like biases.
\newblock \emph{Science}, 356(6334):183--186.

\bibitem[{de~Vassimon~Manela et~al.(2021)de~Vassimon~Manela, Errington, Fisher,
  van Breugel, and Minervini}]{Manela2021StereotypeAS}
Daniel de~Vassimon~Manela, David Errington, Thomas Fisher, Boris van Breugel,
  and Pasquale Minervini. 2021.
\newblock Stereotype and skew: Quantifying gender bias in pre-trained and
  fine-tuned language models.
\newblock In \emph{EACL}.

\bibitem[{Devlin et~al.(2019)Devlin, Chang, Lee, and
  Toutanova}]{devlin2019bert}
Jacob Devlin, Ming-Wei Chang, Kenton Lee, and Kristina Toutanova. 2019.
\newblock \href {https://doi.org/10.18653/v1/N19-1423} {{BERT}: Pre-training of
  deep bidirectional transformers for language understanding}.
\newblock In \emph{Proceedings of the 2019 Conference of the North {A}merican
  Chapter of the Association for Computational Linguistics: Human Language
  Technologies, Volume 1 (Long and Short Papers)}, pages 4171--4186,
  Minneapolis, Minnesota. Association for Computational Linguistics.

\bibitem[{Francis and Kucera(1979)}]{francis1979brown}
W~Nelson Francis and Henry Kucera. 1979.
\newblock Brown corpus manual.
\newblock \emph{Letters to the Editor}, 5(2):7.

\bibitem[{Gonen and Goldberg(2019)}]{gonen-goldberg-2019-lipstick}
Hila Gonen and Yoav Goldberg. 2019.
\newblock \href {https://doi.org/10.18653/v1/N19-1061} {Lipstick on a pig:
  {D}ebiasing methods cover up systematic gender biases in word embeddings but
  do not remove them}.
\newblock In \emph{Proceedings of the 2019 Conference of the North {A}merican
  Chapter of the Association for Computational Linguistics: Human Language
  Technologies, Volume 1 (Long and Short Papers)}, pages 609--614, Minneapolis,
  Minnesota. Association for Computational Linguistics.

\bibitem[{He and Jaeger(2018)}]{he2018overcoming}
Xu~He and Herbert Jaeger. 2018.
\newblock Overcoming catastrophic interference using conceptor-aided
  backpropagation.
\newblock In \emph{International Conference on Learning Representations}.

\bibitem[{Jaeger(2014)}]{jaeger2014controlling}
Herbert Jaeger. 2014.
\newblock Controlling recurrent neural networks by conceptors.
\newblock \emph{arXiv preprint arXiv:1403.3369}.

\bibitem[{Kaneko and Bollegala(2021)}]{kaneko2021debiasing}
Masahiro Kaneko and Danushka Bollegala. 2021.
\newblock Debiasing pre-trained contextualised embeddings.
\newblock In \emph{Proceedings of the 16th Conference of the European Chapter
  of the Association for Computational Linguistics: Main Volume}, pages
  1256--1266.

\bibitem[{Karve et~al.(2019)Karve, Ungar, and Sedoc}]{karve2019conceptor}
Saket Karve, Lyle Ungar, and Jo{\~a}o Sedoc. 2019.
\newblock Conceptor debiasing of word representations evaluated on weat.
\newblock In \emph{Proceedings of the First Workshop on Gender Bias in Natural
  Language Processing}, pages 40--48.

\bibitem[{Kurita et~al.(2019)Kurita, Vyas, Pareek, Black, and
  Tsvetkov}]{kurita-etal-2019-measuring}
Keita Kurita, Nidhi Vyas, Ayush Pareek, Alan~W Black, and Yulia Tsvetkov. 2019.
\newblock \href {https://doi.org/10.18653/v1/W19-3823} {Measuring bias in
  contextualized word representations}.
\newblock In \emph{Proceedings of the First Workshop on Gender Bias in Natural
  Language Processing}, pages 166--172, Florence, Italy. Association for
  Computational Linguistics.

\bibitem[{Lalor et~al.(2022)Lalor, Yang, Smith, Forsgren, and
  Abbasi}]{lalor2022intersectional}
John Lalor, Yi~Yang, Kendall Smith, Nicole Forsgren, and Ahmed Abbasi. 2022.
\newblock \href {https://doi.org/10.18653/v1/2022.naacl-main.263} {Benchmarking
  intersectional biases in {NLP}}.
\newblock In \emph{Proceedings of the 2022 Conference of the North American
  Chapter of the Association for Computational Linguistics: Human Language
  Technologies}, pages 3598--3609, Seattle, United States. Association for
  Computational Linguistics.

\bibitem[{Lhoest et~al.(2021)Lhoest, Villanova~del Moral, Jernite, Thakur, von
  Platen, Patil, Chaumond, Drame, Plu, Tunstall, Davison, {\v{S}}a{\v{s}}ko,
  Chhablani, Malik, Brandeis, Le~Scao, Sanh, Xu, Patry, McMillan-Major, Schmid,
  Gugger, Delangue, Matussi{\`e}re, Debut, Bekman, Cistac, Goehringer, Mustar,
  Lagunas, Rush, and Wolf}]{lhoest-etal-2021-datasets}
Quentin Lhoest, Albert Villanova~del Moral, Yacine Jernite, Abhishek Thakur,
  Patrick von Platen, Suraj Patil, Julien Chaumond, Mariama Drame, Julien Plu,
  Lewis Tunstall, Joe Davison, Mario {\v{S}}a{\v{s}}ko, Gunjan Chhablani,
  Bhavitvya Malik, Simon Brandeis, Teven Le~Scao, Victor Sanh, Canwen Xu,
  Nicolas Patry, Angelina McMillan-Major, Philipp Schmid, Sylvain Gugger,
  Cl{\'e}ment Delangue, Th{\'e}o Matussi{\`e}re, Lysandre Debut, Stas Bekman,
  Pierric Cistac, Thibault Goehringer, Victor Mustar, Fran{\c{c}}ois Lagunas,
  Alexander Rush, and Thomas Wolf. 2021.
\newblock \href {https://doi.org/10.18653/v1/2021.emnlp-demo.21} {Datasets: A
  community library for natural language processing}.
\newblock In \emph{Proceedings of the 2021 Conference on Empirical Methods in
  Natural Language Processing: System Demonstrations}, pages 175--184, Online
  and Punta Cana, Dominican Republic. Association for Computational
  Linguistics.

\bibitem[{Liang et~al.(2020)Liang, Li, Zheng, Lim, Salakhutdinov, and
  Morency}]{liang2020towards}
Paul~Pu Liang, Irene~Mengze Li, Emily Zheng, Yao~Chong Lim, Ruslan
  Salakhutdinov, and Louis-Philippe Morency. 2020.
\newblock Towards debiasing sentence representations.
\newblock In \emph{Proceedings of the 58th Annual Meeting of the Association
  for Computational Linguistics}.

\bibitem[{Liu et~al.(2018)Liu, Sedoc, and Ungar}]{liu2018correcting}
Tianlin Liu, Jo{\~a}o Sedoc, and Lyle Ungar. 2018.
\newblock Correcting the common discourse bias in linear representation of
  sentences using conceptors.
\newblock \emph{arXiv preprint arXiv:1811.11002}.

\bibitem[{Liu et~al.(2019{\natexlab{a}})Liu, Ungar, and
  Sedoc}]{liu2019continual}
Tianlin Liu, Lyle Ungar, and Jo{\~a}o Sedoc. 2019{\natexlab{a}}.
\newblock Continual learning for sentence representations using conceptors.
\newblock In \emph{Proceedings of the 2019 Conference of the North American
  Chapter of the Association for Computational Linguistics: Human Language
  Technologies, Volume 1 (Long and Short Papers)}, pages 3274--3279.

\bibitem[{Liu et~al.(2019{\natexlab{b}})Liu, Ungar, and
  Sedoc}]{liu2019unsupervised}
Tianlin Liu, Lyle Ungar, and Joao Sedoc. 2019{\natexlab{b}}.
\newblock Unsupervised post-processing of word vectors via conceptor negation.
\newblock In \emph{Proceedings of the AAAI Conference on Artificial
  Intelligence}, volume~33, pages 6778--6785.

\bibitem[{May et~al.(2019)May, Wang, Bordia, Bowman, and
  Rudinger}]{may2019seat}
Chandler May, Alex Wang, Shikha Bordia, Samuel Bowman, and Rachel Rudinger.
  2019.
\newblock On measuring social biases in sentence encoders.
\newblock In \emph{Proceedings of the 2019 Conference of the North American
  Chapter of the Association for Computational Linguistics: Human Language
  Technologies, Volume 1 (Long and Short Papers)}, pages 622--628.

\bibitem[{McCoy et~al.(2019)McCoy, Min, and Linzen}]{mccoy2019berts}
R~Thomas McCoy, Junghyun Min, and Tal Linzen. 2019.
\newblock Berts of a feather do not generalize together: Large variability in
  generalization across models with similar test set performance.
\newblock \emph{arXiv preprint arXiv:1911.02969}.

\bibitem[{McInnes et~al.(2018)McInnes, Healy, and Melville}]{mcinnes2018umap}
Leland McInnes, John Healy, and James Melville. 2018.
\newblock Umap: Uniform manifold approximation and projection for dimension
  reduction.
\newblock \emph{arXiv preprint arXiv:1802.03426}.

\bibitem[{Meade et~al.(2022)Meade, Poole-Dayan, and
  Reddy}]{meade_2022_empirical}
Nicholas Meade, Elinor Poole-Dayan, and Siva Reddy. 2022.
\newblock \href {https://doi.org/10.18653/v1/2022.acl-long.132} {An empirical
  survey of the effectiveness of debiasing techniques for pre-trained language
  models}.
\newblock In \emph{Proceedings of the 60th Annual Meeting of the Association
  for Computational Linguistics (Volume 1: Long Papers)}, pages 1878--1898,
  Dublin, Ireland. Association for Computational Linguistics.

\bibitem[{Nadeem et~al.(2021)Nadeem, Bethke, and
  Reddy}]{nadeem-etal-2021-stereoset}
Moin Nadeem, Anna Bethke, and Siva Reddy. 2021.
\newblock \href {https://doi.org/10.18653/v1/2021.acl-long.416} {{S}tereo{S}et:
  Measuring stereotypical bias in pretrained language models}.
\newblock In \emph{Proceedings of the 59th Annual Meeting of the Association
  for Computational Linguistics and the 11th International Joint Conference on
  Natural Language Processing (Volume 1: Long Papers)}, pages 5356--5371,
  Online. Association for Computational Linguistics.

\bibitem[{Nangia et~al.(2020)Nangia, Vania, Bhalerao, and
  Bowman}]{nangia-etal-2020-crows}
Nikita Nangia, Clara Vania, Rasika Bhalerao, and Samuel~R. Bowman. 2020.
\newblock \href {https://doi.org/10.18653/v1/2020.emnlp-main.154}
  {{C}row{S}-pairs: A challenge dataset for measuring social biases in masked
  language models}.
\newblock In \emph{Proceedings of the 2020 Conference on Empirical Methods in
  Natural Language Processing (EMNLP)}, pages 1953--1967, Online. Association
  for Computational Linguistics.

\bibitem[{Peters et~al.(2018)Peters, Neumann, Iyyer, Gardner, Clark, Lee, and
  Zettlemoyer}]{peters-etal-2018-deep}
Matthew~E. Peters, Mark Neumann, Mohit Iyyer, Matt Gardner, Christopher Clark,
  Kenton Lee, and Luke Zettlemoyer. 2018.
\newblock \href {https://doi.org/10.18653/v1/N18-1202} {Deep contextualized
  word representations}.
\newblock In \emph{Proceedings of the 2018 Conference of the North {A}merican
  Chapter of the Association for Computational Linguistics: Human Language
  Technologies, Volume 1 (Long Papers)}, pages 2227--2237, New Orleans,
  Louisiana. Association for Computational Linguistics.

\bibitem[{Radford et~al.(2019)Radford, Wu, Child, Luan, Amodei, and
  Sutskever}]{radford2019language}
Alec Radford, Jeff Wu, Rewon Child, David Luan, Dario Amodei, and Ilya
  Sutskever. 2019.
\newblock Language models are unsupervised multitask learners.

\bibitem[{Ravfogel et~al.(2020)Ravfogel, Elazar, Gonen, Twiton, and
  Goldberg}]{ravfogel-etal-2020-null}
Shauli Ravfogel, Yanai Elazar, Hila Gonen, Michael Twiton, and Yoav Goldberg.
  2020.
\newblock \href {https://doi.org/10.18653/v1/2020.acl-main.647} {Null it out:
  Guarding protected attributes by iterative nullspace projection}.
\newblock In \emph{Proceedings of the 58th Annual Meeting of the Association
  for Computational Linguistics}, pages 7237--7256, Online. Association for
  Computational Linguistics.

\bibitem[{Schick et~al.(2021)Schick, Udupa, and Sch{\"u}tze}]{schick2021self}
Timo Schick, Sahana Udupa, and Hinrich Sch{\"u}tze. 2021.
\newblock Self-diagnosis and self-debiasing: A proposal for reducing
  corpus-based bias in nlp.
\newblock \emph{Transactions of the Association for Computational Linguistics},
  9:1408--1424.

\bibitem[{Sedoc and Ungar(2019)}]{sedoc2019role}
Jo{\~a}o Sedoc and Lyle Ungar. 2019.
\newblock The role of protected class word lists in bias identification of
  contextualized word representations.
\newblock In \emph{Proceedings of the First Workshop on Gender Bias in Natural
  Language Processing}, pages 55--61.

\bibitem[{Socher et~al.(2013)Socher, Perelygin, Wu, Chuang, Manning, Ng, and
  Potts}]{socher2013recursive}
Richard Socher, Alex Perelygin, Jean Wu, Jason Chuang, Christopher~D Manning,
  Andrew~Y Ng, and Christopher Potts. 2013.
\newblock Recursive deep models for semantic compositionality over a sentiment
  treebank.
\newblock In \emph{Proceedings of the 2013 conference on empirical methods in
  natural language processing}, pages 1631--1642.

\bibitem[{Tan and Celis(2019)}]{tan2019assessing}
Yi~Chern Tan and L~Elisa Celis. 2019.
\newblock Assessing social and intersectional biases in contextualized word
  representations.
\newblock \emph{Advances in neural information processing systems}, 32.

\bibitem[{Van~der Maaten and Hinton(2008)}]{van2008visualizing}
Laurens Van~der Maaten and Geoffrey Hinton. 2008.
\newblock Visualizing data using t-sne.
\newblock \emph{Journal of machine learning research}, 9(11).

\bibitem[{Wang et~al.(2018)Wang, Singh, Michael, Hill, Levy, and
  Bowman}]{wang2018glue}
Alex Wang, Amanpreet Singh, Julian Michael, Felix Hill, Omer Levy, and
  Samuel~R. Bowman. 2018.
\newblock {GLUE}: A multi-task benchmark and analysis platform for natural
  language understanding.
\newblock ArXiv preprint 1804.07461.

\bibitem[{Webster et~al.(2020)Webster, Wang, Tenney, Beutel, Pitler, Pavlick,
  Chen, Chi, and Petrov}]{webster2020measuring}
Kellie Webster, Xuezhi Wang, Ian Tenney, Alex Beutel, Emily Pitler, Ellie
  Pavlick, Jilin Chen, Ed~Chi, and Slav Petrov. 2020.
\newblock Measuring and reducing gendered correlations in pre-trained models.
\newblock \emph{arXiv preprint arXiv:2010.06032}.

\bibitem[{Wolf et~al.(2019)Wolf, Debut, Sanh, Chaumond, Delangue, Moi, Cistac,
  Rault, Louf, Funtowicz, and Brew}]{Wolf2019HuggingFacesTS}
Thomas Wolf, Lysandre Debut, Victor Sanh, Julien Chaumond, Clement Delangue,
  Anthony Moi, Pierric Cistac, Tim Rault, R{\'e}mi Louf, Morgan Funtowicz, and
  Jamie Brew. 2019.
\newblock Huggingface's transformers: State-of-the-art natural language
  processing.
\newblock \emph{ArXiv}, abs/1910.03771.

\bibitem[{Xie and Lukasiewicz(2023)}]{xie2023empirical}
Zhongbin Xie and Thomas Lukasiewicz. 2023.
\newblock An empirical analysis of parameter-efficient methods for debiasing
  pre-trained language models.
\newblock \emph{arXiv preprint arXiv:2306.04067}.

\bibitem[{Zhao et~al.(2019)Zhao, Wang, Yatskar, Cotterell, Ordonez, and
  Chang}]{zhao2019gender}
Jieyu Zhao, Tianlu Wang, Mark Yatskar, Ryan Cotterell, Vicente Ordonez, and
  Kai-Wei Chang. 2019.
\newblock Gender bias in contextualized word embeddings.
\newblock In \emph{NAACL (short)}.

\bibitem[{Zhao et~al.(2018)Zhao, Wang, Yatskar, Ordonez, and
  Chang}]{zhao2018gender}
Jieyu Zhao, Tianlu Wang, Mark Yatskar, Vicente Ordonez, and Kai-Wei Chang.
  2018.
\newblock Gender bias in coreference resolution: Evaluation and debiasing
  methods.
\newblock In \emph{NAACL (short)}.

\end{thebibliography}
\bibliographystyle{acl_natbib}

\clearpage
\appendix
\label{sec:appendix}

\section{Attribute Wordlist}

The examples and sources of the attribute wordlists are given below. Due to space limitations, we would only provide up to 50 words for each list. 

\subsection{Pronouns Wordlist}

Words (in total 22): \textit{son, mother, daughter, him, brother, girl, uncle, hers, grandfather, his, boy, her, father, she, sister, man, female, aunt, woman, grandmother, he, male}. 

They are the concatenation of W7\_terms and W8\_terms from WEAT list\footnote{
\url{https://github.com/jsedoc/ConceptorDebias/blob/master/lists/WEAT_lists.py}}.

\subsection{Extended Wordlist}

Words (randomly 50 of 388): \textit{paramour, abbesses, headmistress, stepson, gods, congressman, gents, uncle, hers, wizard, cowgirls, fiancees, adultress, sororal, ladies, sons, uncles, actors, beards, heiress, fellas, salesman, princess, empress, masters, chairwomen, miss, horsewomen, actor, mr., strongwoman, barons, andrology, busboy, prince, hens, womb, masseuse, lady, testosterone, daughter, girl, stateswoman, businessmen, women, fraternities, aunts, boys, abbot, heroine, …}

They are the concatenation of lists: WinoBias extra gendered words\footnote{
\url{https://github.com/uclanlp/corefBias/blob/master/_site/WinoBias/wino/extra_gendered_words.txt}}, 
GN-GloVe male's name\footnote{\url{https://github.com/uclanlp/gn_glove/blob/master/wordlist/female_word_file.txt }}, and 
female's name\footnote{\url{https://github.com/uclanlp/gn_glove/blob/master/wordlist/male_word_file.txt }}. 

\subsection{Propernouns Wordlist}

Words (randomly 50 of 7578): \textit{Broddie, Tony, Tawsha, Emylee, Orelle, Gerrilee, Katusha, Georges, Reine, Hayley, Deloria, Richmond, Wilfrid, Neille, Florie, Riva, Sandro, Cooper, Thom, Pate, Nikoletta, Rodrique, Pat, Chuck, Theressa, Brett, Kaspar, Elric, Storm, Yule, Bubba, Thomasina, Anson, Margery, Abra, Benedict, Cy, Gertrud, Morly, Julina, Melly, Quinta, Paolo, Brynne, Maurene, Alexis, Ramsey, Sianna, Phebe, Alfred, …} 

They are the concatenation of lists: CMU male's name\footnote{\url{ https://www.cs.cmu.edu/Groups/AI/areas/nlp/corpora/names/male.txt}} and 
female's name\footnote{\url{https://www.cs.cmu.edu/Groups/AI/areas/nlp/corpora/names/female.txt}}.

\subsection{Race Wordlist}

Words (in total 8): \textit{africa, african, america, asia, asian, caucasian, china, europe}

\subsection{Outlier Filtering}

Table~\ref{tab:num-word-percentile} shows the number of remaining gender words per percentile after being filtered on UMAP-clustering space.

\begin{table}[h]
    \centering
    \small
    \begin{tabular}{ccccc}
    \hline \multirow{2}*{Percentile} & \multicolumn{4}{c}{ Bias Subspace Type } \\ 
        \cline { 2 - 5 } & pronouns & extended & propernouns & all \\ 
    \hline  0.5-1.0 & 22 & 388  & 7578 & 7988 \\
    \hline  0.4 & 22 & 388  & 5443 & 6902 \\
    \hline  0.3 & 11 & 372  & 4942 & 5140 \\
    \hline  0.2 &  7 & 364  & 3194 & 3289 \\
    \hline  0.1 &  4 &  67  & 1067 & 1087 \\
    \hline
    \end{tabular}
    \caption{\label{tab:num-word-percentile} The number of remaining words per percentile after filtering on UMAP-clustering space. The model is ``bert-base-uncased''.}
\end{table}

\section{Model Checkpoints}

We use the Hugging Face Transformers package~\cite{Wolf2019HuggingFacesTS} in our experiments. The models and checkpoint names are given in Table~\ref{tab:checkpoint}.

\begin{table}[H]
    \centering
    \begin{tabular}{ll}
    \hline Model & Checkpoint \\
    \hline 
    BERT-T & \verb|prajjwal1/bert-tiny| \\ 
    BERT & \verb|bert-base-uncased| \\
    BERT-L & \verb|bert-large-uncased| \\
    GPT2 & \verb|gpt2| \\
    GPT2-L & \verb|gpt2-large| \\
    GPT-J & \verb|EleutherAI/gpt-j-6B| \\
    \hline
    \end{tabular}
    \caption{\label{tab:checkpoint} The package's model and checkpoint name in our experiment.}
\end{table}

\section{Continued Training Details}

The conceptor-intervened model is trained for one epoch by setting \verb|prediction_loss_only| as true and \verb|per_device_train_batch_size| as 8. Following the training procedure in \citet{devlin2019bert}, we train by tasks Masked Language Model (MLM) and Next Sentence Prediction (NSP) simultaneously. The training corpus is the Wikipedia dump from datasets library~\cite{lhoest-etal-2021-datasets}.

\section{GLUE Details}

Before being evaluated on GLUE, each model is trained for three epochs with the following settings: \verb|batch_size| $32$, \verb|maximum_sequence_length| $128$, and \verb|learning_rate| $2\mathrm{e}{-}5$; the same as~\citet{meade_2022_empirical}.

\section{Full Bert-Base-Uncased Model Results\label{app:additional-bert-base}}

\begin{itemize}
    \item Table~\ref{tab:seat-belt-different-corpus} shows the gender debiasing result by different types of the corpus, using the last layer of ``bert-base-uncased'' as a benchmark.
    \item Table~\ref{tab:seat-belt-different-percentiles-on-brown}, and \ref{tab:seat-belt-different-percentiles-on-sst}, \ref{tab:seat-belt-different-percentiles-on-reddit} show the post-processing gender debiasing result of different percentiles of wordlist on three different corpora: Brown, SST, and Reddit, respectively.
    \item Table~\ref{tab:seat-belt-different-layer-conceptor-intervened-brown} and \ref{tab:seat-belt-different-layer-conceptor-intervened-sst} show the post-processing and conceptor-intervened gender debiasing result of each layer on two different corpora: Brown and SST, respectively.
    \item Table~\ref{tab:glue-bert-full} contains GLUE results for the gender debiased model. 
\end{itemize}


\begin{table*}[h]
\centering
\small
\begin{tabular}{lllllllr}
\hline Model & SEAT-6 & SEAT-6b & SEAT-7 & SEAT-7b & SEAT-8 & SEAT-8b & Avg. Abs. \\
\hline 
BERT (``bert-base-uncased'') & 0.931$^*$ & 0.090 & -0.124 & 0.937$^*$ & 0.783$^*$ & 0.858$^*$ & 0.620 \\
\hdashline
(Brown Corpus) & & & & & & & \\
+ Conceptor-12 (pronouns) & \textbf{0.488}$^*$ & -0.091 & -0.331 & \textbf{0.471}$^*$ & 0.783$^*$ & \textbf{0.621}$^*$ & \ColorDown{0.156} \textbf{0.464} \\
+ Conceptor-12 (extended) & \textbf{0.509}$^*$ & -0.109 & -0.406 & \textbf{0.240} & \textbf{0.606}$^*$ & \textbf{0.449}$^*$ & \ColorDown{0.234} \ColorBest{\textbf{0.386}} \\
+ Conceptor-12 (propernouns) & \textbf{0.581}$^*$ & \textbf{-0.053} & -0.258 & \textbf{0.187} & \textbf{0.585}$^*$ & \textbf{0.659}$^*$ & \ColorDown{0.233} \ColorBest{\textbf{0.387}} \\
+ Conceptor-12 (all) & \textbf{0.452}$^*$ & -0.123 & -0.277 & \textbf{0.258} & \textbf{0.662}$^*$ & \textbf{0.607}$^*$ & \ColorDown{0.224} \textbf{0.396} \\
+ Conceptor-12 (or) & \textbf{0.440}$^*$ & \textbf{-0.063} & -0.136 & \textbf{0.251} & \textbf{0.640}$^*$ & \textbf{0.617}$^*$ & \ColorDown{0.262} \ColorBest{\textbf{0.358}} \\

\hdashline
(SST Corpus) & & & & & & & \\
+ Conceptor-12 (pronouns) & \textbf{0.627}$^*$ & -0.104 & -0.416 & \textbf{0.520}$^*$ & \textbf{0.636}$^*$ & \textbf{0.628}$^*$ & \ColorDown{0.132} \textbf{0.488} \\
+ Conceptor-12 (extended) & \textbf{0.626}$^*$ & \textbf{-0.068} & -0.365 & \textbf{0.280} & \textbf{0.556}$^*$ & \textbf{0.429} & \ColorDown{0.233} \ColorBest{\textbf{0.387}} \\
+ Conceptor-12 (propernouns) & \textbf{0.680}$^*$ & \textbf{-0.050} & -0.405 & \textbf{0.614}$^*$ & \textbf{0.585}$^*$ & \textbf{0.790}$^*$ & \ColorDown{0.099} \textbf{0.521} \\
+ Conceptor-12 (all) & \textbf{0.624}$^*$ & -0.093 & -0.480 & \textbf{0.442}$^*$ & \textbf{0.538}$^*$ & \textbf{0.663}$^*$ & \ColorDown{0.147} \textbf{0.473} \\
+ Conceptor-12 (or) & \textbf{0.606}$^*$ & -0.100 & -0.447 & \textbf{0.337} & \textbf{0.428} & \textbf{0.575}$^*$ & \ColorDown{0.204} \textbf{0.416} \\

\hdashline
(Reddit Corpus) & & & & & & & \\
+ Conceptor-12 (pronouns) & \textbf{0.619}$^*$ & -0.092 & -0.235 & \textbf{0.816}$^*$ & \textbf{0.756}$^*$ & 0.962$^*$ & \ColorDown{0.040} \textbf{0.580} \\
+ Conceptor-12 (extended) & \textbf{0.630}$^*$ & \textbf{-0.061} & -0.157 & \textbf{0.676}$^*$ & \textbf{0.711}$^*$ & \textbf{0.806}$^*$ & \ColorDown{0.113} \textbf{0.507} \\
+ Conceptor-12 (propernouns) & \textbf{0.792}$^*$ & 0.151 & \textbf{0.068} & 0.964$^*$ & \textbf{0.765}$^*$ & 0.934$^*$ & \ColorDown{0.008} \textbf{0.612} \\
+ Conceptor-12 (all) & \textbf{0.613}$^*$ & \textbf{-0.010} & \textbf{0.004} & \textbf{0.803}$^*$ & \textbf{0.735}$^*$ & 0.917$^*$ & \ColorDown{0.106} \textbf{0.514} \\
+ Conceptor-12 (or) & \textbf{0.593}$^*$ & \textbf{0.004} & \textbf{0.092} & \textbf{0.838}$^*$ & \textbf{0.652}$^*$ & 0.961$^*$ & \ColorDown{0.097} \textbf{0.523} \\

\hdashline
+ CDA & \textbf{0.846}$^*$ & 0.186 & -0.278 & 1.342$^*$ & 0.831$^*$ & \textbf{0.849}$^*$ & \ColorUp{0.120} 0.722 \\
+ \textsc{Dropout} & \textbf{1.136}$^*$ & 0.317 & 0.138 & 1.179$^*$ & 0.879$^*$ & 0.939$^*$ & \ColorUp{0.144}  0.765 \\
+ INLP & \textbf{0.317} & -0.354 & -0.258 & \textbf{0.105} & \textbf{0.187} & \textbf{-0.004} & \ColorDown{0.416} \textbf{0.204} \\
+ \textsc{SentenceDebias} & \textbf{0.350} & -0.298 & -0.626 & \textbf{0.458}$^*$ & \textbf{0.413} & \textbf{0.462}$^*$ & \ColorDown{0.186} \textbf{0.434}\\
\hline
\end{tabular}
\caption{\label{tab:seat-belt-different-corpus} SEAT effect size of gender debising. The impact of \textit{different corpora} on \textit{bert-base-uncased} models. Effect sizes closer to 0 are indicative of less biased sentence representations (\textbf{bolded value}). Statistically significant effect sizes at $p<0.01$ are denoted by *. Note that the ``conceptor-$X$ (subspace)'' indicates the conceptor negation matrix is generated by the $X$-layer of the language model in combinations with the subspace of the specific attribute wordlist. The top-3 best performance is colored in orange.
}
\end{table*}


\begin{table*}[thb!]
\centering
\small 
\begin{tabular}{lllllllr}
\hline Model & SEAT-6 & SEAT-6b & SEAT-7 & SEAT-7b & SEAT-8 & SEAT-8b & Avg. Abs. \\
\hline 
BERT (``bert-base-uncased'') & 0.931$^*$ & 0.090 & -0.124 & 0.937$^*$ & 0.783$^*$ & 0.858$^*$ & 0.620 \\
\hdashline
(Wordlist Percentile 0.5-1.0)\\
+ Conceptor-12 (pronouns) & \textbf{0.488}$^*$ & -0.091 & -0.331 & \textbf{0.471}$^*$ & 0.783$^*$ & \textbf{0.621}$^*$ & \ColorDown{0.156} \textbf{0.464} \\
+ Conceptor-12 (extended) & \textbf{0.509}$^*$ & -0.109 & -0.406 & \textbf{0.240} & \textbf{0.606}$^*$ & \textbf{0.449}$^*$ & \ColorDown{0.234} \textbf{0.386} \\
+ Conceptor-12 (propernouns) & \textbf{0.581}$^*$ & \textbf{-0.053} & -0.258 & \textbf{0.187} & \textbf{0.585}$^*$ & \textbf{0.659}$^*$ & \ColorDown{0.233} \textbf{0.387} \\
+ Conceptor-12 (all) & \textbf{0.452}$^*$ & -0.123 & -0.277 & \textbf{0.258} & \textbf{0.662}$^*$ & \textbf{0.607}$^*$ & \ColorDown{0.224} \textbf{0.396} \\
+ Conceptor-12 (or) & \textbf{0.440}$^*$ & \textbf{-0.063} & -0.136 & \textbf{0.251} & \textbf{0.640}$^*$ & \textbf{0.617}$^*$ & \ColorDown{0.262} \textbf{0.358} \\
\hdashline
(Wordlist Percentile 0.4)\\
+ Conceptor-12 (pronouns) & \textbf{0.483}$^*$ & -0.095 & -0.385 & \textbf{0.435}$^*$ & \textbf{0.776}$^*$ & \textbf{0.609}$^*$ & \ColorDown{0.156} \textbf{0.464} \\
+ Conceptor-12 (extended) & \textbf{0.509}$^*$ & -0.110 & -0.407 & \textbf{0.239} & \textbf{0.603}$^*$ & \textbf{0.447}$^*$ & \ColorDown{0.234} \textbf{0.386} \\
+ Conceptor-12 (propernouns) & \textbf{0.451}$^*$ & -0.188 & -0.505 & \textbf{-0.122} & \textbf{0.399} & \textbf{0.264} & \ColorDown{0.298} \ColorBest{\textbf{0.322}} \\
+ Conceptor-12 (all) & \textbf{0.466}$^*$ & -0.112 & -0.260 & \textbf{0.267} & \textbf{0.697}$^*$ & \textbf{0.617}$^*$ & \ColorDown{0.217} \textbf{0.403} \\
+ Conceptor-12 (or) & \textbf{0.388} & \textbf{-0.078} & -0.292 & \textbf{0.179} & \textbf{0.594}$^*$ & \textbf{0.335} & \ColorDown{0.309} \ColorBest{\textbf{0.311}} \\
\hdashline
(Wordlist Percentile 0.3)\\
+ Conceptor-12 (pronouns) & \textbf{0.487}$^*$ & \textbf{-0.016} & -0.351 & \textbf{0.398}$^*$ & 0.807$^*$ & \textbf{0.776}$^*$ & \ColorDown{0.148} \textbf{0.472} \\
+ Conceptor-12 (extended) & \textbf{0.509}$^*$ & -0.109 & -0.410 & \textbf{0.228} & \textbf{0.604}$^*$ & \textbf{0.453}$^*$ & \ColorDown{0.235} \textbf{0.385} \\
+ Conceptor-12 (propernouns) & \textbf{0.495}$^*$ & -0.168 & -0.478 & \textbf{-0.083} & \textbf{0.456}$^*$ & \textbf{0.315} & \ColorDown{0.288} \ColorBest{\textbf{0.332}} \\
+ Conceptor-12 (all) & \textbf{0.348} & -0.236 & -0.520 & \textbf{-0.019} & \textbf{0.506}$^*$ & \textbf{0.361} & \ColorDown{0.288} \ColorBest{\textbf{0.332}} \\
+ Conceptor-12 (or) & \textbf{0.407} & \textbf{-0.022} & -0.247 & \textbf{0.331} & \textbf{0.677}$^*$ & \textbf{0.483}$^*$ & \ColorDown{0.259} \textbf{0.361} \\
\hdashline
(Wordlist Percentile 0.2)\\
+ Conceptor-12 (pronouns) & \textbf{0.570}$^*$ & \textbf{0.035} & -0.378 & \textbf{0.334} & \textbf{0.708}$^*$ & \textbf{0.768}$^*$ & \ColorDown{0.154} \textbf{0.466} \\
+ Conceptor-12 (extended) & \textbf{0.508}$^*$ & -0.109 & -0.416 & \textbf{0.219} & \textbf{0.602}$^*$ & \textbf{0.450}$^*$ & \ColorDown{0.236} \textbf{0.384} \\
+ Conceptor-12 (propernouns) & \textbf{0.548}$^*$ & -0.157 & -0.397 & \textbf{0.270} & \textbf{0.483}$^*$ & \textbf{0.366} & \ColorDown{0.250} \textbf{0.370} \\
+ Conceptor-12 (all) & \textbf{0.357} & -0.235 & -0.598 & \textbf{0.110} & \textbf{0.455}$^*$ & \textbf{0.383} & \ColorDown{0.264} \textbf{0.356} \\
+ Conceptor-12 (or) & \textbf{0.476}$^*$ & \textbf{-0.063} & -0.385 & \textbf{0.296} & \textbf{0.558}$^*$ & \textbf{0.500}$^*$ & \ColorDown{0.240} \textbf{0.380} \\
\hdashline
(Wordlist Percentile 0.1)\\
+ Conceptor-12 (pronouns) & \textbf{0.800}$^*$ & 0.204 & -0.314 & \textbf{0.273} & \textbf{0.764}$^*$ & 0.965$^*$ & \ColorDown{0.067} \textbf{0.553} \\
+ Conceptor-12 (extended) & \textbf{0.869}$^*$ & 0.162 & -0.265 & \textbf{0.266} & 0.861$^*$ & \textbf{0.635}$^*$ & \ColorDown{0.110} \textbf{0.510} \\
+ Conceptor-12 (propernouns) & \textbf{0.613}$^*$ & \textbf{-0.084} & -0.582 & \textbf{0.190} & \textbf{0.579}$^*$ & \textbf{0.740}$^*$ & \ColorDown{0.155} \textbf{0.465} \\
+ Conceptor-12 (all) & \textbf{0.603}$^*$ & -0.102 & -0.612 & \textbf{0.182} & \textbf{0.566}$^*$ & \textbf{0.712}$^*$ & \ColorDown{0.157} \textbf{0.463} \\
+ Conceptor-12 (or) & \textbf{0.614}$^*$ & 0.197 & -0.401 & \textbf{-0.132} & \textbf{0.624}$^*$ & \textbf{0.699}$^*$ & \ColorDown{0.176} \textbf{0.444} \\

\hdashline
+ CDA & \textbf{0.846}$^*$ & 0.186 & -0.278 & 1.342$^*$ & 0.831$^*$ & \textbf{0.849}$^*$ & \ColorUp{0.120} 0.722 \\
+ \textsc{Dropout} & \textbf{1.136}$^*$ & 0.317 & 0.138 & 1.179$^*$ & 0.879$^*$ & 0.939$^*$ & \ColorUp{0.144}  0.765 \\
+ INLP & \textbf{0.317} & -0.354 & -0.258 & \textbf{0.105} & \textbf{0.187} & \textbf{-0.004} & \ColorDown{0.416} \textbf{0.204} \\
+ \textsc{SentenceDebias} & \textbf{0.350} & -0.298 & -0.626 & \textbf{0.458}$^*$ & \textbf{0.413} & \textbf{0.462}$^*$ & \ColorDown{0.186} \textbf{0.434}\\
\hline
\end{tabular}
\caption{\label{tab:seat-belt-different-percentiles-on-brown} SEAT effect size of gender debising. The impact of \textit{different percentiles of wordlist} (using \textit{UMAP} clustering) on \textit{Brown} Corpus, \textit{bert-base-uncased} models. The top-3 best performance is colored in orange.
}
\end{table*}

\begin{table*}[thb!]
\centering
\small 
\begin{tabular}{lllllllr}
\hline Model & SEAT-6 & SEAT-6b & SEAT-7 & SEAT-7b & SEAT-8 & SEAT-8b & Avg. Abs. \\
\hline

BERT (``bert-base-uncased'') & 0.931$^*$ & 0.090 & -0.124 & 0.937$^*$ & 0.783$^*$ & 0.858$^*$ & 0.620 \\
\hdashline

(Wordlist Percentile 0.5-1.0)\\
+ Conceptor-12 (pronouns) & \textbf{0.627}$^*$ & -0.104 & -0.416 & \textbf{0.520}$^*$ & \textbf{0.636}$^*$ & \textbf{0.628}$^*$ & \ColorDown{0.132} \textbf{0.488} \\
+ Conceptor-12 (extended) & \textbf{0.688}$^*$ & \textbf{0.024} & -0.293 & \textbf{0.138} & \textbf{0.559}$^*$ & \textbf{0.375} & \ColorDown{0.274} \ColorBest{\textbf{0.346}} \\
+ Conceptor-12 (propernouns) & \textbf{0.680}$^*$ & \textbf{-0.050} & -0.405 & \textbf{0.614}$^*$ & \textbf{0.585}$^*$ & \textbf{0.790}$^*$ & \ColorDown{0.099} \textbf{0.521} \\
+ Conceptor-12 (all) & \textbf{0.624}$^*$ & -0.093 & -0.480 & \textbf{0.442}$^*$ & \textbf{0.538}$^*$ & \textbf{0.663}$^*$ & \ColorDown{0.147} \textbf{0.473} \\
+ Conceptor-12 (or) & \textbf{0.619}$^*$ & \textbf{-0.069} & -0.428 & \textbf{0.280} & \textbf{0.414} & \textbf{0.539}$^*$ & \ColorDown{0.229} \textbf{0.391} \\
\hdashline
(Wordlist Percentile 0.4)\\
+ Conceptor-12 (pronouns) & \textbf{0.619}$^*$ & -0.113 & -0.526 & \textbf{0.449}$^*$ & \textbf{0.606}$^*$ & \textbf{0.584}$^*$ & \ColorDown{0.137} \textbf{0.483} \\
+ Conceptor-12 (extended) & \textbf{0.688}$^*$ & \textbf{0.024} & -0.293 & \textbf{0.138} & \textbf{0.559}$^*$ & \textbf{0.375} & \ColorDown{0.274} \ColorBest{\textbf{0.346}} \\
+ Conceptor-12 (propernouns) & \textbf{0.704}$^*$ & \textbf{-0.086} & -0.227 & \textbf{0.590}$^*$ & \textbf{0.682}$^*$ & \textbf{0.716}$^*$ & \ColorDown{0.119} \textbf{0.501} \\
+ Conceptor-12 (all) & \textbf{0.622}$^*$ & \textbf{-0.087} & -0.508 & \textbf{0.277} & \textbf{0.519}$^*$ & \textbf{0.578}$^*$ & \ColorDown{0.188} \textbf{0.432} \\
+ Conceptor-12 (or) & \textbf{0.646}$^*$ & \textbf{-0.034} & -0.427 & \textbf{0.200} & \textbf{0.438}$^*$ & \textbf{0.401} & \ColorDown{0.262} \ColorBest{\textbf{0.358}} \\
\hdashline
(Wordlist Percentile 0.3)\\
+ Conceptor-12 (pronouns) & \textbf{0.550}$^*$ & \textbf{-0.035} & -0.396 & \textbf{0.344} & \textbf{0.682}$^*$ & \textbf{0.744}$^*$ & \ColorDown{0.161} \textbf{0.459} \\
+ Conceptor-12 (extended) & \textbf{0.687}$^*$ & \textbf{0.023} & -0.299 & \textbf{0.129} & \textbf{0.559}$^*$ & \textbf{0.382} & \ColorDown{0.273} \ColorBest{\textbf{0.347}} \\
+ Conceptor-12 (propernouns) & \textbf{0.706}$^*$ & \textbf{-0.088} & -0.230 & \textbf{0.602}$^*$ & \textbf{0.683}$^*$ & \textbf{0.720}$^*$ & \ColorDown{0.115} \textbf{0.505} \\
+ Conceptor-12 (all) & \textbf{0.652}$^*$ & -0.117 & -0.378 & \textbf{0.504}$^*$ & \textbf{0.536}$^*$ & \textbf{0.657}$^*$ & \ColorDown{0.146} \textbf{0.474} \\
+ Conceptor-12 (or) & \textbf{0.595}$^*$ & \textbf{0.027} & -0.375 & \textbf{0.171} & \textbf{0.519}$^*$ & \textbf{0.600}$^*$ & \ColorDown{0.239} \textbf{0.381} \\
\hdashline
(Wordlist Percentile 0.2)\\
+ Conceptor-12 (pronouns) & \textbf{0.730}$^*$ & 0.090 & \textbf{-0.110} & \textbf{0.523}$^*$ & \textbf{0.714}$^*$ & \textbf{0.758}$^*$ & \ColorDown{0.132} \textbf{0.488} \\
+ Conceptor-12 (extended) & \textbf{0.687}$^*$ & \textbf{0.023} & -0.299 & \textbf{0.129} & \textbf{0.559}$^*$ & \textbf{0.382} & \ColorDown{0.273} \ColorBest{\textbf{0.347}} \\
+ Conceptor-12 (propernouns) & \textbf{0.755}$^*$ & \textbf{-0.057} & -0.346 & \textbf{0.584}$^*$ & \textbf{0.659}$^*$ & \textbf{0.733}$^*$ & \ColorDown{0.098} \textbf{0.522} \\
+ Conceptor-12 (all) & \textbf{0.699}$^*$ & -0.094 & -0.492 & \textbf{0.456}$^*$ & \textbf{0.620}$^*$ & \textbf{0.688}$^*$ & \ColorDown{0.112} \textbf{0.508} \\
+ Conceptor-12 (or) & \textbf{0.579}$^*$ & \textbf{-0.004} & -0.261 & \textbf{0.270} & \textbf{0.503}$^*$ & \textbf{0.634}$^*$ & \ColorDown{0.245} \textbf{0.375} \\
\hdashline
(Wordlist Percentile 0.1)\\
+ Conceptor-12 (pronouns) & \textbf{0.903}$^*$ & 0.198 & -0.464 & \textbf{0.030} & \textbf{0.434} & \textbf{0.492}$^*$ & \ColorDown{0.200} \textbf{0.420} \\
+ Conceptor-12 (extended) & \textbf{0.631}$^*$ & -0.107 & \textbf{-0.012} & 1.008$^*$ & \textbf{0.615}$^*$ & \textbf{0.794}$^*$ & \ColorDown{0.092} \textbf{0.528} \\
+ Conceptor-12 (propernouns) & \textbf{0.655}$^*$ & -0.130 & -0.166 & \textbf{0.895}$^*$ & \textbf{0.641}$^*$ & 0.877$^*$ & \ColorDown{0.059} \textbf{0.561} \\
+ Conceptor-12 (all) & \textbf{0.597}$^*$ & -0.164 & -0.223 & \textbf{0.791}$^*$ & \textbf{0.698}$^*$ & 0.909$^*$ & \ColorDown{0.056} \textbf{0.564} \\
+ Conceptor-12 (or) & \textbf{0.542}$^*$ & \textbf{-0.039} & -0.184 & \textbf{0.589}$^*$ & \textbf{0.457}$^*$ & 0.929$^*$ & \ColorDown{0.163} \textbf{0.457} \\

\hdashline
+ CDA & \textbf{0.846}$^*$ & 0.186 & -0.278 & 1.342$^*$ & 0.831$^*$ & \textbf{0.849}$^*$ & \ColorUp{0.120} 0.722 \\
+ \textsc{Dropout} & \textbf{1.136}$^*$ & 0.317 & 0.138 & 1.179$^*$ & 0.879$^*$ & 0.939$^*$ & \ColorUp{0.144}  0.765 \\
+ INLP & \textbf{0.317} & -0.354 & -0.258 & \textbf{0.105} & \textbf{0.187} & \textbf{-0.004} & \ColorDown{0.416} \textbf{0.204} \\
+ \textsc{SentenceDebias} & \textbf{0.350} & -0.298 & -0.626 & \textbf{0.458}$^*$ & \textbf{0.413} & \textbf{0.462}$^*$ & \ColorDown{0.186} \textbf{0.434}\\
\hline
\end{tabular}
\caption{\label{tab:seat-belt-different-percentiles-on-sst} SEAT effect size of gender debising. The impact of \textit{different percentiles of wordlist} (using \textit{UMAP} clustering) on \textit{SST} Corpus, \textit{bert-base-uncased} models. The top-3 best performance is colored in orange.
}
\end{table*}

\begin{table*}[thb!]
\centering
\small
\begin{tabular}{lllllllr}
\hline Model & SEAT-6 & SEAT-6b & SEAT-7 & SEAT-7b & SEAT-8 & SEAT-8b & Avg. Abs. \\
\hline

BERT (``bert-base-uncased'') & 0.931$^*$ & 0.090 & -0.124 & 0.937$^*$ & 0.783$^*$ & 0.858$^*$ & 0.620 \\
\hdashline
(Wordlist Percentile 0.5-1.0)\\
+ Conceptor-12 (pronouns) & \textbf{0.619}$^*$ & -0.092 & -0.235 & \textbf{0.816}$^*$ & \textbf{0.756}$^*$ & 0.962$^*$ & \ColorDown{0.040} \textbf{0.580} \\
+ Conceptor-12 (extended) & \textbf{0.630}$^*$ & \textbf{-0.061} & -0.157 & \textbf{0.676}$^*$ & \textbf{0.711}$^*$ & \textbf{0.806}$^*$ & \ColorDown{0.113} \textbf{0.507} \\
+ Conceptor-12 (propernouns) & \textbf{0.792}$^*$ & 0.151 & \textbf{0.068} & 0.964$^*$ & \textbf{0.765}$^*$ & 0.934$^*$ & \ColorDown{0.008} \textbf{0.612} \\
+ Conceptor-12 (all) & \textbf{0.613}$^*$ & \textbf{-0.010} & \textbf{0.004} & \textbf{0.803}$^*$ & \textbf{0.735}$^*$ & 0.917$^*$ & \ColorDown{0.106} \textbf{0.514} \\
+ Conceptor-12 (or) & \textbf{0.593}$^*$ & \textbf{0.004} & \textbf{0.092} & \textbf{0.838}$^*$ & \textbf{0.652}$^*$ & 0.961$^*$ & \ColorDown{0.097} \textbf{0.523} \\
\hdashline
(Wordlist Percentile 0.4)\\
+ Conceptor-12 (pronouns) & \textbf{0.618}$^*$ & -0.095 & -0.297 & \textbf{0.769}$^*$ & \textbf{0.728}$^*$ & 0.944$^*$ & \ColorDown{0.045} \textbf{0.575} \\
+ Conceptor-12 (extended) & \textbf{0.732}$^*$ & \textbf{0.047} & \textbf{-0.121} & \textbf{0.601}$^*$ & \textbf{0.748}$^*$ & \textbf{0.719}$^*$ & \ColorDown{0.125} \ColorBest{\textbf{0.495}} \\
+ Conceptor-12 (propernouns) & \textbf{0.691}$^*$ & \textbf{-0.063} & 0.315 & 1.028$^*$ & \textbf{0.667}$^*$ & \textbf{0.635}$^*$ & \ColorDown{0.054} \textbf{0.566} \\
+ Conceptor-12 (all) & \textbf{0.651}$^*$ & \textbf{-0.010} & \textbf{-0.004} & \textbf{0.786}$^*$ & \textbf{0.732}$^*$ & 0.940$^*$ & \ColorDown{0.100} \textbf{0.520} \\
+ Conceptor-12 (or) & \textbf{0.526}$^*$ & \textbf{-0.057} & \textbf{0.054} & \textbf{0.596}$^*$ & \textbf{0.655}$^*$ & \textbf{0.724}$^*$ & \ColorDown{0.185} \ColorBest{\textbf{0.435}} \\
\hdashline
(Wordlist Percentile 0.3)\\
+ Conceptor-12 (pronouns) & \textbf{0.596}$^*$ & \textbf{-0.068} & -0.286 & \textbf{0.772}$^*$ & 0.790$^*$ & 0.983$^*$ & \ColorDown{0.038} \textbf{0.582} \\
+ Conceptor-12 (extended) & \textbf{0.740}$^*$ & \textbf{0.043} & \textbf{-0.119} & \textbf{0.593}$^*$ & \textbf{0.758}$^*$ & \textbf{0.725}$^*$ & \ColorDown{0.124} \textbf{0.496} \\
+ Conceptor-12 (propernouns) & \textbf{0.825}$^*$ & \textbf{0.007} & 0.512$^*$ & 1.180$^*$ & \textbf{0.761}$^*$ & \textbf{0.705}$^*$ & \ColorUp{0.045} \textbf{0.665} \\
+ Conceptor-12 (all) & \textbf{0.689}$^*$ & \textbf{0.007} & 0.180 & \textbf{0.924}$^*$ & \textbf{0.627}$^*$ & \textbf{0.653}$^*$ & \ColorDown{0.107} \textbf{0.513} \\
+ Conceptor-12 (or) & \textbf{0.612}$^*$ & \textbf{-0.018} & \textbf{0.046} & \textbf{0.745}$^*$ & \textbf{0.778}$^*$ & 0.916$^*$ & \ColorDown{0.101} \textbf{0.519} \\
\hdashline
(Wordlist Percentile 0.2)\\
+ Conceptor-12 (pronouns) & \textbf{0.801}$^*$ & 0.104 & \textbf{-0.072} & \textbf{0.757}$^*$ & 0.873$^*$ & 0.997$^*$ & \ColorDown{0.019} \textbf{0.601} \\
+ Conceptor-12 (extended) & \textbf{0.740}$^*$ & \textbf{0.043} & \textbf{-0.119} & \textbf{0.593}$^*$ & \textbf{0.758}$^*$ & \textbf{0.725}$^*$ & \ColorDown{0.124} \textbf{0.496} \\
+ Conceptor-12 (propernouns) & \textbf{0.837}$^*$ & \textbf{0.037} & 0.532$^*$ & 1.170$^*$ & 0.785$^*$ & \textbf{0.722}$^*$ & \ColorUp{0.060} \textbf{0.680} \\
+ Conceptor-12 (all) & \textbf{0.694}$^*$ & \textbf{-0.042} & 0.356 & 1.044$^*$ & \textbf{0.608}$^*$ & \textbf{0.554}$^*$ & \ColorDown{0.070} \textbf{0.550} \\
+ Conceptor-12 (or) & \textbf{0.665}$^*$ & 0.096 & 0.130 & \textbf{0.607}$^*$ & 0.852$^*$ & \textbf{0.842}$^*$ & \ColorDown{0.088} \textbf{0.532} \\
\hdashline
(Wordlist Percentile 0.1)\\
+ Conceptor-12 (pronouns) & 0.949$^*$ & 0.121 & -0.472 & \textbf{0.102} & \textbf{0.568}$^*$ & \textbf{0.707}$^*$ & \ColorDown{0.134} \ColorBest{\textbf{0.486}} \\
+ Conceptor-12 (extended) & \textbf{0.736}$^*$ & \textbf{-0.035} & -0.291 & \textbf{0.780}$^*$ & 0.812$^*$ & 1.095$^*$ & \ColorUp{0.005} \textbf{0.625} \\
+ Conceptor-12 (propernouns) & 0.948$^*$ & 0.107 & \textbf{-0.094} & 0.949$^*$ & 0.783$^*$ & \textbf{0.842}$^*$ & --\qquad \textbf{0.620} \\
+ Conceptor-12 (all) & 0.949$^*$ & 0.105 & \textbf{-0.061} & \textbf{0.899}$^*$ & \textbf{0.782}$^*$ & \textbf{0.846}$^*$ & \ColorDown{0.013} \textbf{0.607} \\
+ Conceptor-12 (or) & 0.936$^*$ & 0.130 & -0.579 & \textbf{0.098} & \textbf{0.591}$^*$ & 0.914$^*$ & \ColorDown{0.079} \textbf{0.541} \\

\hdashline
+ CDA & \textbf{0.846}$^*$ & 0.186 & -0.278 & 1.342$^*$ & 0.831$^*$ & \textbf{0.849}$^*$ & \ColorUp{0.120} 0.722 \\
+ \textsc{Dropout} & \textbf{1.136}$^*$ & 0.317 & 0.138 & 1.179$^*$ & 0.879$^*$ & 0.939$^*$ & \ColorUp{0.144}  0.765 \\
+ INLP & \textbf{0.317} & -0.354 & -0.258 & \textbf{0.105} & \textbf{0.187} & \textbf{-0.004} & \ColorDown{0.416} \textbf{0.204} \\
+ \textsc{SentenceDebias} & \textbf{0.350} & -0.298 & -0.626 & \textbf{0.458}$^*$ & \textbf{0.413} & \textbf{0.462}$^*$ & \ColorDown{0.186} \textbf{0.434}\\
\hline
\end{tabular}
\caption{\label{tab:seat-belt-different-percentiles-on-reddit} SEAT effect size of gender debising. The impact of \textit{different percentiles of wordlist} (using \textit{UMAP} clustering) on \textit{Reddit} Corpus, \textit{bert-base-uncased} models. The top-3 best performance is colored in orange.
}
\end{table*}


\begin{table*}[thb!]
\centering
\small 
\begin{tabular}{lllllllr}
\hline Model & SEAT-6 & SEAT-6b & SEAT-7 & SEAT-7b & SEAT-8 & SEAT-8b & Avg. Abs. \\
\hline

(Layer 0)\\
BERT (``bert-base-uncased'') & 0.921$^*$ & 0.194 & 0.251 & -0.172 & -0.110 & 0.366 & 0.336 \\
+ Conceptor-0 (or) & \textbf{0.147} & \textbf{-0.087} & -0.266 & -0.653 & -0.405 & \textbf{-0.324} & \ColorDown{0.022} \textbf{0.314} \\
+ Conceptor-Intervened & \textbf{0.147} & \textbf{-0.087} & -0.266 & -0.653 & -0.405 & \textbf{-0.324} & \ColorDown{0.022} \textbf{0.314} \\
\hdashline
(Layer 1)\\
BERT (``bert-base-uncased'') & 1.245$^*$ & 0.292 & 0.469$^*$ & 1.101$^*$ & 0.110 & 1.261$^*$ & 0.746 \\
+ Conceptor-1 (or) & \textbf{0.473}$^*$ & \textbf{0.205} & \textbf{-0.210} & \textbf{-0.093} & \textbf{-0.095} & \textbf{0.396} & \ColorDown{0.501} \textbf{0.245} \\
+ Conceptor-Intervened & \textbf{0.241} & \textbf{-0.038} & \textbf{-0.274} & \textbf{0.291} & -0.751 & \textbf{-0.107} & \ColorDown{0.462} \textbf{0.284} \\
\hdashline
(Layer 2)\\
BERT (``bert-base-uncased'') & 1.149$^*$ & 0.216 & 0.431$^*$ & 1.021$^*$ & 0.474$^*$ & 1.231$^*$ & 0.754 \\
+ Conceptor-2 (or) & \textbf{0.180} & \textbf{-0.047} & -0.450 & \textbf{-0.105} & \textbf{0.133} & \textbf{0.133} & \ColorDown{0.579} \textbf{0.175} \\
+ Conceptor-Intervened & \textbf{-0.108} & \textbf{0.115} & -0.965 & 1.388$^*$ & -1.146 & \textbf{0.329} & \ColorDown{0.079} \textbf{0.675} \\
\hdashline
(Layer 3)\\
BERT (``bert-base-uncased'') & 1.186$^*$ & 0.214 & 0.152 & 0.770$^*$ & 0.262 & 1.049$^*$ & 0.606 \\
+ Conceptor-3 (or) & \textbf{0.404} & \textbf{-0.046} & -0.675 & \textbf{0.102} & -0.325 & \textbf{-0.024} & \ColorDown{0.343} \textbf{0.263} \\
+ Conceptor-Intervened & \textbf{0.158} & \textbf{0.081} & -0.959 & 1.348$^*$ & -1.093 & \textbf{0.409} & \ColorUp{0.069} \textbf{0.675} \\
\hdashline
(Layer 4)\\
BERT (``bert-base-uncased'') & 0.975$^*$ & 0.106 & 0.552$^*$ & 0.890$^*$ & 0.542$^*$ & 0.724$^*$ & 0.632 \\
+ Conceptor-4 (or) & \textbf{0.597}$^*$ & \textbf{0.068} & \textbf{-0.249} & \textbf{0.251} & \textbf{-0.016} & \textbf{-0.315} & \ColorDown{0.383} \textbf{0.249} \\
+ Conceptor-Intervened & \textbf{0.060} & 0.121 & -0.986 & 1.676$^*$ & -0.840 & 0.943$^*$ & \ColorUp{0.139} \textbf{0.771} \\
\hdashline
(Layer 5)\\
BERT (``bert-base-uncased'') & 1.002$^*$ & 0.184 & 0.628$^*$ & 0.914$^*$ & 0.376 & 1.053$^*$ & 0.693 \\
+ Conceptor-5 (or) & \textbf{0.634}$^*$ & \textbf{0.064} & \textbf{0.118} & \textbf{0.225} & \textbf{-0.160} & \textbf{0.429} & \ColorDown{0.421} \textbf{0.272} \\
+ Conceptor-Intervened & \textbf{-0.046} & \textbf{0.043} & -1.038 & 1.378$^*$ & -0.790 & \textbf{0.659}$^*$ & \ColorDown{0.034} \textbf{0.659} \\
\hdashline
(Layer 6)\\
BERT (``bert-base-uncased'') & 0.753$^*$ & 0.118 & 0.539$^*$ & 1.048$^*$ & 0.597$^*$ & 1.042$^*$ & 0.683 \\
+ Conceptor-6 (or) & \textbf{0.327} & \textbf{0.041} & \textbf{0.176} & \textbf{0.104} & \textbf{0.150} & \textbf{0.174} & \ColorDown{0.521} \textbf{0.162} \\
+ Conceptor-Intervened & \textbf{-0.210} & \textbf{0.004} & -0.965 & 1.242$^*$ & -0.739 & \textbf{0.475}$^*$ & \ColorDown{0.077} \textbf{0.606} \\
\hdashline
(Layer 7)\\
BERT (``bert-base-uncased'') & 0.719$^*$ & 0.155 & 0.341 & 0.935$^*$ & 0.562$^*$ & 0.721$^*$ & 0.572 \\
+ Conceptor-7 (or) & \textbf{0.235} & \textbf{-0.064} & \textbf{-0.038} & \textbf{0.206} & \textbf{0.173} & \textbf{0.223} & \ColorDown{0.416} \textbf{0.156} \\
+ Conceptor-Intervened & \textbf{-0.246} & \textbf{-0.082} & -0.821 & 1.112$^*$ & -0.671 & \textbf{0.248} & \ColorDown{0.042} \textbf{0.530} \\
\hdashline
(Layer 8)\\
BERT (``bert-base-uncased'') & 0.983$^*$ & 0.163 & 0.313 & 1.157$^*$ & 0.766$^*$ & 0.789$^*$ & 0.695 \\
+ Conceptor-8 (or) & \textbf{0.235} & \textbf{0.005} & \textbf{-0.136} & \textbf{0.389} & \textbf{0.379} & \textbf{0.135} & \ColorDown{0.482} \textbf{0.213} \\
+ Conceptor-Intervened & \textbf{-0.125} & -0.193 & -0.940 & \textbf{0.796}$^*$ & \textbf{-0.606} & \textbf{0.084} & \ColorDown{0.238} \textbf{0.457} \\
\hdashline
(Layer 9)\\
BERT (``bert-base-uncased'') & 0.922$^*$ & 0.224 & 0.503$^*$ & 1.293$^*$ & 0.780$^*$ & 0.996$^*$ & 0.786 \\
+ Conceptor-9 (or) & \textbf{0.234} & \textbf{0.019} & \textbf{-0.005} & \textbf{0.485}$^*$ & \textbf{0.694}$^*$ & \textbf{0.686}$^*$ & \ColorDown{0.432} \textbf{0.354} \\
+ Conceptor-Intervened & \textbf{-0.151} & -0.246 & -0.599 & \textbf{0.836}$^*$ & \textbf{-0.455} & \textbf{-0.095} & \ColorDown{0.389} \textbf{0.397} \\
\hdashline
(Layer 10)\\
BERT (``bert-base-uncased'') & 0.686$^*$ & 0.082 & 0.226 & 0.894$^*$ & 0.904$^*$ & 0.965$^*$ & 0.626 \\
+ Conceptor-10 (or) & \textbf{0.294} & -0.091 & \textbf{-0.153} & \textbf{0.078} & \textbf{0.703}$^*$ & \textbf{0.545}$^*$ & \ColorDown{0.315} \textbf{0.311} \\
+ Conceptor-Intervened & \textbf{-0.253} & -0.298 & -0.569 & \textbf{0.753}$^*$ & \textbf{-0.462} & \textbf{-0.099} & \ColorDown{0.221} \textbf{0.405} \\
\hdashline
(Layer 11)\\
BERT (``bert-base-uncased'') & 0.665$^*$ & -0.015 & -0.344 & 0.602$^*$ & 0.919$^*$ & 0.891$^*$ & 0.573 \\
+ Conceptor-11 (or) & \textbf{0.197} & -0.114 & -0.399 & \textbf{-0.157} & \textbf{0.557}$^*$ & \textbf{0.277} & \ColorDown{0.289} \textbf{0.284} \\
+ Conceptor-Intervened & \textbf{-0.314} & -0.269 & -0.635 & 0.769$^*$ & \textbf{-0.430} & \textbf{0.096} & \ColorDown{0.154} \textbf{0.419} \\
\hdashline
(Layer 12)\\
BERT (``bert-base-uncased'') & 0.931$^*$ & 0.090 & -0.124 & 0.937$^*$ & 0.783$^*$ & 0.858$^*$ & 0.620 \\
+ Conceptor-12 (or) & \textbf{0.388} & \textbf{-0.078} & -0.292 & \textbf{0.179} & \textbf{0.594}$^*$ & \textbf{0.335} & \ColorDown{0.309} \textbf{0.311} \\
+ Conceptor-Intervened & \textbf{-0.334} & -0.117 & -0.698 & \textbf{0.459}$^*$ & \textbf{-0.230} & \textbf{0.178} & \ColorDown{0.284} \textbf{0.336} \\
\hdashline

+ CDA & \textbf{0.846}$^*$ & 0.186 & -0.278 & 1.342$^*$ & 0.831$^*$ & \textbf{0.849}$^*$ & \ColorUp{0.120} 0.722 \\
+ \textsc{Dropout} & \textbf{1.136}$^*$ & 0.317 & 0.138 & 1.179$^*$ & 0.879$^*$ & 0.939$^*$ & \ColorUp{0.144}  0.765 \\
+ INLP & \textbf{0.317} & -0.354 & -0.258 & \textbf{0.105} & \textbf{0.187} & \textbf{-0.004} & \ColorDown{0.416} \textbf{0.204} \\
+ \textsc{SentenceDebias} & \textbf{0.350} & -0.298 & -0.626 & \textbf{0.458}$^*$ & \textbf{0.413} & \textbf{0.462}$^*$ & \ColorDown{0.186} \textbf{0.434}\\

\hline
\end{tabular}
\caption{\label{tab:seat-belt-different-layer-conceptor-intervened-brown} SEAT fffect size of gender debising from CI-BERT, Type I.
The conceptor-intervened performance of \textit{different layer's conceptors} on \textit{SST} Corpus, \textit{bert-base-uncased} models. The setting is ``brown-0.4-or''. The layer(s) of CI-BERT that outperform the conceptor post-processing of the same layer(s) are colored in orange.
}
\end{table*}

\begin{table*}[h]
\centering
\small
\begin{tabular}{lllllllr}
\hline Model & SEAT-6 & SEAT-6b & SEAT-7 & SEAT-7b & SEAT-8 & SEAT-8b & Avg. Abs. \\
\hline

(Layer 0)\\
BERT (``bert-base-uncased'') & 0.921$^*$ & 0.194 & 0.251 & -0.172 & -0.110 & 0.366 & 0.336 \\
+ Conceptor-0 (extended) & \textbf{0.497}$^*$ & \textbf{-0.095} & -0.412 & -0.760 & \textbf{-0.001} & \textbf{-0.276} & \ColorUp{0.004} \textbf{0.340} \\
+ Conceptor-Intervened & \textbf{0.497}$^*$ & \textbf{-0.095} & -0.412 & -0.760 & \textbf{-0.001} & \textbf{-0.276} & \ColorUp{0.004} \textbf{0.340} \\
\hdashline
(Layer 1)\\
BERT (``bert-base-uncased'') & 1.245$^*$ & 0.292 & 0.469$^*$ & 1.101$^*$ & 0.110 & 1.261$^*$ & 0.746 \\
+ Conceptor-1 (extended) & \textbf{0.897}$^*$ & \textbf{0.156} & \textbf{-0.084} & \textbf{0.208} & \textbf{0.099} & \textbf{0.558}$^*$ & \ColorDown{0.412} \textbf{0.334} \\
+ Conceptor-Intervened & \textbf{0.813}$^*$ & \textbf{0.029} & -0.961 & \textbf{-0.513} & -0.211 & \textbf{-0.292} & \ColorDown{0.276} \textbf{0.470} \\
\hdashline
(Layer 2)\\
BERT (``bert-base-uncased'') & 1.149$^*$ & 0.216 & 0.431$^*$ & 1.021$^*$ & 0.474$^*$ & 1.231$^*$ & 0.754 \\
+ Conceptor-2 (extended) & \textbf{0.542}$^*$ & \textbf{0.059} & \textbf{-0.146} & \textbf{0.112} & 0.515$^*$ & \textbf{0.428} & \ColorDown{0.454} \textbf{0.300} \\
+ Conceptor-Intervened & \textbf{0.366} & \textbf{0.088} & -1.342 & \textbf{-0.249} & \textbf{-0.408} & \textbf{-0.634} & \ColorDown{0.239} \textbf{0.515} \\
\hdashline
(Layer 3)\\
BERT (``bert-base-uncased'') & 1.186$^*$ & 0.214 & 0.152 & 0.770$^*$ & 0.262 & 1.049$^*$ & 0.606 \\
+ Conceptor-3 (extended) & \textbf{0.849}$^*$ & \textbf{-0.034} & -0.516 & \textbf{0.154} & \textbf{0.205} & \textbf{0.356} & \ColorDown{0.254} \textbf{0.352} \\
+ Conceptor-Intervened & \textbf{0.329} & \textbf{0.048} & -1.240 & \textbf{-0.520} & -0.429 & \textbf{-0.499} & \ColorDown{0.095} \textbf{0.511} \\
\hdashline
(Layer 4)\\
BERT (``bert-base-uncased'') & 0.975$^*$ & 0.106 & 0.552$^*$ & 0.890$^*$ & 0.542$^*$ & 0.724$^*$ & 0.632 \\
+ Conceptor-4 (extended) & \textbf{0.789}$^*$ & 0.109 & \textbf{-0.014} & \textbf{0.254} & \textbf{0.515}$^*$ & \textbf{0.009} & \ColorDown{0.350} \textbf{0.282} \\
+ Conceptor-Intervened & \textbf{0.248} & \textbf{0.068} & -1.367 & \textbf{-0.040} & \textbf{-0.401} & \textbf{-0.100} & \ColorDown{0.261} \textbf{0.371} \\
\hdashline
(Layer 5)\\
BERT (``bert-base-uncased'') & 1.002$^*$ & 0.184 & 0.628$^*$ & 0.914$^*$ & 0.376 & 1.053$^*$ & 0.693 \\
+ Conceptor-5 (extended) & \textbf{0.695}$^*$ & \textbf{0.122} & \textbf{-0.007} & \textbf{0.075} & \textbf{0.158} & \textbf{0.472}$^*$ & \ColorDown{0.438} \textbf{0.255} \\
+ Conceptor-Intervened & \textbf{0.105} & \textbf{0.066} & -1.096 & \textbf{-0.170} & \textbf{-0.173} & \textbf{0.036} & \ColorDown{0.419} \textbf{0.274} \\
\hdashline
(Layer 6)\\
BERT (``bert-base-uncased'') & 0.753$^*$ & 0.118 & 0.539$^*$ & 1.048$^*$ & 0.597$^*$ & 1.042$^*$ & 0.683 \\
+ Conceptor-6 (extended) & \textbf{0.372} & \textbf{0.084} & \textbf{0.150} & \textbf{0.033} & \textbf{0.467}$^*$ & \textbf{0.209} & \ColorDown{0.464} \textbf{0.219} \\
+ Conceptor-Intervened & \textbf{0.004} & \textbf{-0.023} & -0.866 & \textbf{-0.312} & \textbf{-0.234} & \textbf{-0.106} & \ColorDown{0.425} \textbf{0.258} \\
\hdashline
(Layer 7)\\
BERT (``bert-base-uncased'') & 0.719$^*$ & 0.155 & 0.341 & 0.935$^*$ & 0.562$^*$ & 0.721$^*$ & 0.572 \\
+ Conceptor-7 (extended) & \textbf{0.451}$^*$ & \textbf{0.082} & \textbf{0.051} & \textbf{0.196} & \textbf{0.326} & \textbf{0.185} & \ColorDown{0.357} \textbf{0.215} \\
+ Conceptor-Intervened & \textbf{0.041} & \textbf{-0.066} & -0.697 & \textbf{-0.509} & \textbf{-0.381} & \textbf{-0.015} & \ColorDown{0.287} \textbf{0.285} \\
\hdashline
(Layer 8)\\
BERT (``bert-base-uncased'') & 0.983$^*$ & 0.163 & 0.313 & 1.157$^*$ & 0.766$^*$ & 0.789$^*$ & 0.695 \\
+ Conceptor-8 (extended) & \textbf{0.597}$^*$ & \textbf{0.051} & \textbf{-0.023} & \textbf{0.639}$^*$ & \textbf{0.503}$^*$ & \textbf{0.200} & \ColorDown{0.359} \textbf{0.336} \\
+ Conceptor-Intervened & \textbf{0.110} & \textbf{-0.095} & -0.392 & \textbf{-0.702} & \textbf{-0.287} & \textbf{0.190} & \ColorDown{0.399} \ColorBest{\textbf{0.296}} \\
\hdashline
(Layer 9)\\
BERT (``bert-base-uncased'') & 0.922$^*$ & 0.224 & 0.503$^*$ & 1.293$^*$ & 0.780$^*$ & 0.996$^*$ & 0.786 \\
+ Conceptor-9 (extended) & \textbf{0.597}$^*$ & \textbf{0.146} & \textbf{0.333} & \textbf{0.903}$^*$ & \textbf{0.764}$^*$ & \textbf{0.722}$^*$ & \ColorDown{0.208} \textbf{0.578} \\
+ Conceptor-Intervened & \textbf{0.148} & \textbf{-0.024} & \textbf{0.162} & \textbf{-0.669} & \textbf{0.224} & \textbf{0.487}$^*$ & \ColorDown{0.500} \ColorBest{\textbf{0.286}} \\
\hdashline
(Layer 10)\\
BERT (``bert-base-uncased'') & 0.686$^*$ & 0.082 & 0.226 & 0.894$^*$ & 0.904$^*$ & 0.965$^*$ & 0.626 \\
+ Conceptor-10 (extended) & \textbf{0.639}$^*$ & 0.099 & \textbf{-0.034} & \textbf{0.044} & \textbf{0.605}$^*$ & \textbf{0.322} & \ColorDown{0.335} \textbf{0.291} \\
+ Conceptor-Intervened & \textbf{0.557}$^*$ & -0.165 & \textbf{-0.149} & -1.046 & \textbf{0.142} & \textbf{0.522}$^*$ & \ColorDown{0.196} \textbf{0.430} \\
\hdashline
(Layer 11)\\
BERT (``bert-base-uncased'') & 0.665$^*$ & -0.015 & -0.344 & 0.602$^*$ & 0.919$^*$ & 0.891$^*$ & 0.573 \\
+ Conceptor-11 (extended) & \textbf{0.565}$^*$ & -0.045 & -0.511 & \textbf{-0.406} & \textbf{0.523}$^*$ & \textbf{0.198} & \ColorDown{0.198} \textbf{0.375} \\
+ Conceptor-Intervened & \textbf{0.602}$^*$ & -0.189 & \textbf{0.143} & -1.219 & \textbf{-0.006} & \textbf{0.205} & \ColorDown{0.179} \textbf{0.394} \\
\hdashline
(Layer 12)\\
BERT (``bert-base-uncased'') & 0.931$^*$ & 0.090 & -0.124 & 0.937$^*$ & 0.783$^*$ & 0.858$^*$ & 0.620 \\
+ Conceptor-12 (extended) & \textbf{0.688}$^*$ & \textbf{0.024} & -0.293 & \textbf{0.138} & \textbf{0.559}$^*$ & \textbf{0.375} & \ColorDown{0.274} \textbf{0.346} \\
+ Conceptor-Intervened & \textbf{0.384} & -0.261 & 0.144 & -1.256 & \textbf{-0.148} & \textbf{0.398} & \ColorDown{0.188} \textbf{0.432} \\
\hdashline

+ CDA & \textbf{0.846}$^*$ & 0.186 & -0.278 & 1.342$^*$ & 0.831$^*$ & \textbf{0.849}$^*$ & \ColorUp{0.120} 0.722 \\
+ \textsc{Dropout} & \textbf{1.136}$^*$ & 0.317 & 0.138 & 1.179$^*$ & 0.879$^*$ & 0.939$^*$ & \ColorUp{0.144}  0.765 \\
+ INLP & \textbf{0.317} & -0.354 & -0.258 & \textbf{0.105} & \textbf{0.187} & \textbf{-0.004} & \ColorDown{0.416} \textbf{0.204} \\
+ \textsc{SentenceDebias} & \textbf{0.350} & -0.298 & -0.626 & \textbf{0.458}$^*$ & \textbf{0.413} & \textbf{0.462}$^*$ & \ColorDown{0.186} \textbf{0.434}\\

\hline
\end{tabular}
\caption{\label{tab:seat-belt-different-layer-conceptor-intervened-sst} SEAT effect size of gender debising from CI-BERT, Type I. The conceptor-intervened performance of \textit{different layer's conceptors} on \textit{SST} Corpus, \textit{bert-base-uncased} models. The setting is ``sst-0.9-extended''. The layer(s) of CI-BERT that outperform the conceptor post-processing of the same layer(s) are colored in orange.
}
\end{table*}


\begin{table*}[]
    \centering
    \small
    \begin{tabular}{lrrrrrrrrrrr}
\hline Model & CoLA & MNLI & MRPC & QNLI & QQP & RTE & SST & STS-B & WNLI & Average \\
\hline BERT  & 55.89 & 84.50 & 88.59 & 91.38 & 91.03 & 63.54 & 92.58 & 88.51 & 43.66 & 77.74 \\
+ Conceptor P.P & 57.54 & 84.66 & 89.30 & 91.03 & 91.05 & 65.34 & 92.66 & 89.07 & 54.93 & \ColorUpCR{1.77} 79.51 \\ 
+ Conceptor C.T. & 47.06 & 83.46 & 87.20 & 90.73 & 90.97 & 58.98 & 91.67 & 88.21 & 52.11 & \ColorDownCR{1.03} 76.71 \\
\hdashline
+ CDA & 55.90 & 84.73 & 88.76 & 91.36 & 91.01 & 66.31 & 92.43 & 89.14 & 38.03 & \ColorDownCR{0.22} 77.52 \\
+ \textsc{Dropout} & 49.83 & 84.67 & 88.20 & 91.27 & 90.36 & 64.02 & 92.58 & 88.47 & 37.09 & \ColorDownCR{1.46} 76.28 \\
+ INLP & 56.06 & 84.81 & 88.61 & 91.34 & 90.92 & 64.98 & 92.51 & 88.70 & 32.86 & \ColorDownCR{0.99} 76.76 \\
+ \textsc{SentenceDebias} & 56.41 & 84.80 & 88.70 & 91.48 & 90.98 & 63.06 & 92.32 & 88.45 & 44.13 & \ColorUpCR{0.07} 77.81 \\
\hline
\end{tabular}
    \caption{GLUE validation set results for gender debiased BERT model. We use the F1 score for MRPC, the Spearman correlation for STS-B, and Matthew's correlation for CoLA. For all other tasks, we report accuracy. All scores are averaged among three runs. The model is ``bert-base-uncased''. The top-3 best performance is colored in orange. }
    \label{tab:glue-bert-full}
\end{table*}

\section{Full Bert-Tiny Model Results\label{app:additional-bert-tiny}}

\begin{itemize}
    \item Table~\ref{tab:seat-bert-tiny-different-percentiles-on-brown}, \ref{tab:seat-bert-tiny-different-percentiles-on-sst}, and  \ref{tab:seat-bert-tiny-different-percentiles-on-reddit} show the post-processing gender debiasing result of different percentiles of wordlist on three different corpora: Brown, SST, and Reddit, respectively.
    \item Table~\ref{tab:seat-bert-tiny-different-layer-conceptor-intervened} shows the post-processing and conceptor-intervened gender debiasing result of each layer on the SST corpus.
\end{itemize}

\begin{table*}[thb!]
\centering
\small 
\begin{tabular}{lllllllr}
\hline Model & SEAT-6 & SEAT-6b & SEAT-7 & SEAT-7b & SEAT-8 & SEAT-8b & Avg. Abs. \\
\hline BERT-T & 1.735$^*$ & 0.797$^*$ & 1.294$^*$ & 1.243$^*$ & 0.837$^*$ & 1.293$^*$ & 1.200 \\
\hdashline

(Wordlist Percentile 1.0)\\
+ Conceptor-2 (pronouns) & \textbf{1.657}$^*$ & \textbf{0.449}$^*$ & \textbf{1.185}$^*$ & \textbf{0.936}$^*$ & \textbf{0.453}$^*$ & \textbf{0.833}$^*$ & \ColorDown{0.281} \textbf{0.919} \\
+ Conceptor-2 (extended) & \textbf{1.570}$^*$ & \textbf{0.353} & \textbf{1.094}$^*$ & \textbf{0.991}$^*$ & \textbf{0.176} & \textbf{0.775}$^*$ & \ColorDown{0.373} \textbf{0.827} \\
+ Conceptor-2 (propernouns) & \textbf{1.641}$^*$ & \textbf{0.655}$^*$ & \textbf{1.142}$^*$ & \textbf{1.121}$^*$ & \textbf{0.203} & \textbf{0.781}$^*$ & \ColorDown{0.276} \textbf{0.924} \\
+ Conceptor-2 (all) & \textbf{1.587}$^*$ & \textbf{0.377}$^*$ & \textbf{1.188}$^*$ & \textbf{1.077}$^*$ & \textbf{0.128} & \textbf{0.735}$^*$ & \ColorDown{0.351} \textbf{0.849} \\
+ Conceptor-2 (or) & \textbf{1.464}$^*$ & \textbf{0.257} & \textbf{1.005}$^*$ & \textbf{0.944}$^*$ & \textbf{-0.114} & \textbf{0.503}$^*$ & \ColorDown{0.486} \ColorBest{\textbf{0.714}} \\
\hdashline
(Wordlist Percentile 0.5-0.9)\\
+ Conceptor-2 (pronouns) & \textbf{1.657}$^*$ & \textbf{0.449}$^*$ & \textbf{1.185}$^*$ & \textbf{0.936}$^*$ & \textbf{0.453}$^*$ & \textbf{0.833}$^*$ & \ColorDown{0.281} \textbf{0.919} \\
+ Conceptor-2 (extended) & \textbf{1.296}$^*$ & \textbf{0.255} & \textbf{1.014}$^*$ & \textbf{1.194}$^*$ & \textbf{-0.274} & \textbf{0.502}$^*$ & \ColorDown{0.444} \textbf{0.756} \\
+ Conceptor-2 (propernouns) & \textbf{1.641}$^*$ & \textbf{0.655}$^*$ & \textbf{1.142}$^*$ & \textbf{1.121}$^*$ & \textbf{0.203} & \textbf{0.781}$^*$ & \ColorDown{0.276} \textbf{0.924} \\
+ Conceptor-2 (all) & \textbf{1.587}$^*$ & \textbf{0.377}$^*$ & \textbf{1.188}$^*$ & \textbf{1.077}$^*$ & \textbf{0.128} & \textbf{0.735}$^*$ & \ColorDown{0.351} \textbf{0.849} \\
+ Conceptor-2 (or) & \textbf{1.323}$^*$ & \textbf{0.186} & \textbf{0.947}$^*$ & \textbf{1.027}$^*$ & \textbf{-0.289} & \textbf{0.403} & \ColorDown{0.504} \ColorBest{\textbf{0.696}} \\
\hdashline
(Wordlist Percentile 0.4)\\
+ Conceptor-2 (pronouns) & \textbf{1.657}$^*$ & \textbf{0.443}$^*$ & \textbf{1.181}$^*$ & \textbf{0.937}$^*$ & \textbf{0.448}$^*$ & \textbf{0.821}$^*$ & \ColorDown{0.286} \textbf{0.914} \\
+ Conceptor-2 (extended) & \textbf{1.294}$^*$ & \textbf{0.254} & \textbf{1.014}$^*$ & \textbf{1.194}$^*$ & \textbf{-0.274} & \textbf{0.502}$^*$ & \ColorDown{0.445} \textbf{0.755} \\
+ Conceptor-2 (propernouns) & \textbf{1.589}$^*$ & \textbf{0.722}$^*$ & \textbf{1.130}$^*$ & \textbf{1.084}$^*$ & \textbf{0.494}$^*$ & \textbf{0.991}$^*$ & \ColorDown{0.198} \textbf{1.002} \\
+ Conceptor-2 (all) & \textbf{1.585}$^*$ & \textbf{0.396}$^*$ & \textbf{1.192}$^*$ & \textbf{1.067}$^*$ & \textbf{0.159} & \textbf{0.726}$^*$ & \ColorDown{0.346} \textbf{0.854} \\
+ Conceptor-2 (or) & \textbf{1.278}$^*$ & \textbf{0.233} & \textbf{0.852}$^*$ & \textbf{0.910}$^*$ & \textbf{-0.265} & \textbf{0.346} & \ColorDown{0.553} \ColorBest{\textbf{0.647}} \\
\hdashline
(Wordlist Percentile 0.3)\\
+ Conceptor-2 (pronouns) & \textbf{1.691}$^*$ & \textbf{0.573}$^*$ & \textbf{1.227}$^*$ & \textbf{1.158}$^*$ & \textbf{0.573}$^*$ & \textbf{1.009}$^*$ & \ColorDown{0.161} \textbf{1.039} \\
+ Conceptor-2 (extended) & \textbf{1.295}$^*$ & \textbf{0.260} & \textbf{1.010}$^*$ & \textbf{1.192}$^*$ & \textbf{-0.286} & \textbf{0.490}$^*$ & \ColorDown{0.444} \textbf{0.756} \\
+ Conceptor-2 (propernouns) & \textbf{1.597}$^*$ & \textbf{0.746}$^*$ & \textbf{1.162}$^*$ & \textbf{1.147}$^*$ & \textbf{0.551}$^*$ & \textbf{1.034}$^*$ & \ColorDown{0.160} \textbf{1.040} \\
+ Conceptor-2 (all) & \textbf{1.536}$^*$ & \textbf{0.436}$^*$ & \textbf{1.143}$^*$ & \textbf{1.181}$^*$ & \textbf{0.140} & \textbf{0.849}$^*$ & \ColorDown{0.319} \textbf{0.881} \\
+ Conceptor-2 (or) & \textbf{1.277}$^*$ & \textbf{0.235} & \textbf{1.055}$^*$ & \textbf{1.168}$^*$ & \textbf{-0.090} & \textbf{0.542}$^*$ & \ColorDown{0.472} \textbf{0.728} \\
\hdashline
(Wordlist Percentile 0.2)\\
+ Conceptor-2 (pronouns) & \textbf{1.656}$^*$ & \textbf{0.543}$^*$ & \textbf{1.253}$^*$ & \textbf{1.175}$^*$ & \textbf{0.569}$^*$ & \textbf{1.038}$^*$ & \ColorDown{0.161} \textbf{1.039} \\
+ Conceptor-2 (extended) & \textbf{1.296}$^*$ & \textbf{0.260} & \textbf{1.011}$^*$ & \textbf{1.189}$^*$ & \textbf{-0.290} & \textbf{0.478}$^*$ & \ColorDown{0.446} \textbf{0.754} \\
+ Conceptor-2 (propernouns) & \textbf{1.577}$^*$ & \textbf{0.723}$^*$ & \textbf{1.231}$^*$ & \textbf{1.193}$^*$ & \textbf{0.541}$^*$ & \textbf{1.067}$^*$ & \ColorDown{0.145} \textbf{1.055} \\
+ Conceptor-2 (all) & \textbf{1.490}$^*$ & \textbf{0.341} & \textbf{1.116}$^*$ & \textbf{1.182}$^*$ & \textbf{-0.018} & \textbf{0.849}$^*$ & \ColorDown{0.367} \textbf{0.833} \\
+ Conceptor-2 (or) & \textbf{1.178}$^*$ & \textbf{0.141} & \textbf{1.062}$^*$ & \textbf{1.087}$^*$ & \textbf{-0.252} & \textbf{0.502}$^*$ & \ColorDown{0.496} \textbf{0.704} \\
\hdashline
(Wordlist Percentile 0.1)\\
+ Conceptor-2 (pronouns) & \textbf{1.677}$^*$ & \textbf{0.643}$^*$ & 1.317$^*$ & 1.342$^*$ & \textbf{0.696}$^*$ & \textbf{1.191}$^*$ & \ColorDown{0.056} \textbf{1.144} \\
+ Conceptor-2 (extended) & \textbf{1.547}$^*$ & \textbf{0.700}$^*$ & 1.305$^*$ & 1.333$^*$ & \textbf{0.464}$^*$ & \textbf{0.997}$^*$ & \ColorDown{0.142} \textbf{1.058} \\
+ Conceptor-2 (propernouns) & \textbf{1.722}$^*$ & 0.836$^*$ & \textbf{1.256}$^*$ & \textbf{1.213}$^*$ & 0.956$^*$ & 1.316$^*$ & \ColorUp{0.017} 1.217 \\
+ Conceptor-2 (all) & 1.771$^*$ & 0.882$^*$ & \textbf{1.189}$^*$ & \textbf{1.160}$^*$ & 0.996$^*$ & \textbf{1.277}$^*$ & \ColorUp{0.012} 1.212 \\
+ Conceptor-2 (or) & \textbf{1.579}$^*$ & \textbf{0.560}$^*$ & \textbf{1.278}$^*$ & 1.301$^*$ & \textbf{0.422} & \textbf{0.881}$^*$ & \ColorDown{0.196} \textbf{1.004} \\

\hline
\end{tabular}
\caption{\label{tab:seat-bert-tiny-different-percentiles-on-brown} SEAT effect size of gender debising. 
The impact of \textit{different percentiles of wordlist} (using \textit{UMAP} clustering) on \textit{Brown} Corpus, \textit{bert-tiny} models. The top-3 best performance is colored in orange. 
}
\end{table*}

\begin{table*}[thb!]
\centering
\small 
\begin{tabular}{lllllllr}
\hline Model & SEAT-6 & SEAT-6b & SEAT-7 & SEAT-7b & SEAT-8 & SEAT-8b & Avg. Abs. \\
\hline BERT-T & 1.735$^*$ & 0.797$^*$ & 1.294$^*$ & 1.243$^*$ & 0.837$^*$ & 1.293$^*$ & 1.200 \\
\hdashline

\hdashline
(Wordlist Percentile 1.0)\\
+ Conceptor-2 (pronouns) & \textbf{1.703}$^*$ & \textbf{0.403}$^*$ & \textbf{0.958}$^*$ & \textbf{0.706}$^*$ & \textbf{0.254} & \textbf{0.679}$^*$ & \ColorDown{0.416} \textbf{0.784} \\
+ Conceptor-2 (extended) & \textbf{1.608}$^*$ & \textbf{0.473}$^*$ & \textbf{0.870}$^*$ & \textbf{1.118}$^*$ & \textbf{-0.209} & \textbf{0.732}$^*$ & \ColorDown{0.365} \textbf{0.835} \\
+ Conceptor-2 (propernouns) & \textbf{1.704}$^*$ & \textbf{0.582}$^*$ & \textbf{1.012}$^*$ & \textbf{1.111}$^*$ & \textbf{-0.069} & \textbf{0.730}$^*$ & \ColorDown{0.332} \textbf{0.868} \\
+ Conceptor-2 (all) & \textbf{1.680}$^*$ & \textbf{0.377}$^*$ & \textbf{1.028}$^*$ & \textbf{1.047}$^*$ & \textbf{-0.175} & \textbf{0.669}$^*$ & \ColorDown{0.371} \textbf{0.829} \\
+ Conceptor-2 (or) & \textbf{1.489}$^*$ & \textbf{0.163} & \textbf{0.539}$^*$ & \textbf{0.937}$^*$ & \textbf{-0.612} & \textbf{0.314} & \ColorDown{0.524} \ColorBest{\textbf{0.676}} \\
\hdashline
(Wordlist Percentile 0.5-0.9)\\
+ Conceptor-2 (pronouns) & \textbf{1.703}$^*$ & \textbf{0.403}$^*$ & \textbf{0.958}$^*$ & \textbf{0.706}$^*$ & \textbf{0.254} & \textbf{0.679}$^*$ & \ColorDown{0.416} \textbf{0.784} \\
+ Conceptor-2 (extended) & \textbf{1.647}$^*$ & \textbf{0.391}$^*$ & \textbf{0.806}$^*$ & \textbf{0.815}$^*$ & \textbf{-0.136} & \textbf{0.637}$^*$ & \ColorDown{0.461} \textbf{0.739} \\
+ Conceptor-2 (propernouns) & \textbf{1.704}$^*$ & \textbf{0.582}$^*$ & \textbf{1.012}$^*$ & \textbf{1.111}$^*$ & \textbf{-0.069} & \textbf{0.730}$^*$ & \ColorDown{0.332} \textbf{0.868} \\
+ Conceptor-2 (all) & \textbf{1.680}$^*$ & \textbf{0.377}$^*$ & \textbf{1.028}$^*$ & \textbf{1.047}$^*$ & \textbf{-0.175} & \textbf{0.669}$^*$ & \ColorDown{0.371} \textbf{0.829} \\
+ Conceptor-2 (or) & \textbf{1.542}$^*$ & \textbf{0.148} & \textbf{0.486}$^*$ & \textbf{0.806}$^*$ & \textbf{-0.549} & \textbf{0.245} & \ColorDown{0.571} \ColorBest{\textbf{0.629}} \\ 
\hdashline
(Wordlist Percentile 0.4)\\
+ Conceptor-2 (pronouns) & \textbf{1.703}$^*$ & \textbf{0.375}$^*$ & \textbf{0.999}$^*$ & \textbf{0.720}$^*$ & \textbf{0.235} & \textbf{0.669}$^*$ & \ColorDown{0.416} \textbf{0.784} \\
+ Conceptor-2 (extended) & \textbf{1.608}$^*$ & \textbf{0.473}$^*$ & \textbf{0.870}$^*$ & \textbf{1.118}$^*$ & \textbf{-0.209} & \textbf{0.732}$^*$ & \ColorDown{0.365} \textbf{0.835} \\
+ Conceptor-2 (propernouns) & 1.765$^*$ & 0.954$^*$ & \textbf{1.027}$^*$ & \textbf{1.036}$^*$ & \textbf{0.457}$^*$ & \textbf{1.028}$^*$ & \ColorDown{0.156} \textbf{1.044} \\
+ Conceptor-2 (all) & \textbf{1.694}$^*$ & \textbf{0.444}$^*$ & \textbf{1.057}$^*$ & \textbf{1.054}$^*$ & \textbf{-0.090} & \textbf{0.709}$^*$ & \ColorDown{0.359} \textbf{0.841} \\
+ Conceptor-2 (or) & \textbf{1.587}$^*$ & \textbf{0.344} & \textbf{0.512}$^*$ & \textbf{0.881}$^*$ & \textbf{-0.407} & \textbf{0.489}$^*$ & \ColorDown{0.497} \ColorBest{\textbf{0.703}} \\
\hdashline
(Wordlist Percentile 0.3)\\
+ Conceptor-2 (pronouns) & 1.786$^*$ & 0.843$^*$ & \textbf{1.121}$^*$ & \textbf{1.028}$^*$ & \textbf{0.617}$^*$ & \textbf{1.064}$^*$ & \ColorDown{0.124} \textbf{1.076} \\
+ Conceptor-2 (extended) & \textbf{1.610}$^*$ & \textbf{0.479}$^*$ & \textbf{0.866}$^*$ & \textbf{1.119}$^*$ & \textbf{-0.215} & \textbf{0.727}$^*$ & \ColorDown{0.364} \textbf{0.836} \\
+ Conceptor-2 (propernouns) & 1.749$^*$ & 0.945$^*$ & \textbf{1.039}$^*$ & \textbf{1.067}$^*$ & \textbf{0.474}$^*$ & \textbf{1.041}$^*$ & \ColorDown{0.148} \textbf{1.052} \\
+ Conceptor-2 (all) & 1.751$^*$ & 0.813$^*$ & \textbf{1.102}$^*$ & \textbf{1.063}$^*$ & \textbf{0.522}$^*$ & \textbf{1.092}$^*$ & \ColorDown{0.143} \textbf{1.057} \\
+ Conceptor-2 (or) & \textbf{1.652}$^*$ & \textbf{0.414}$^*$ & \textbf{0.784}$^*$ & \textbf{1.047}$^*$ & \textbf{-0.170} & \textbf{0.725}$^*$ & \ColorDown{0.401} \textbf{0.799} \\
\hdashline
(Wordlist Percentile 0.2)\\
+ Conceptor-2 (pronouns) & 1.862$^*$ & 0.983$^*$ & \textbf{1.178}$^*$ & \textbf{0.982}$^*$ & \textbf{0.737}$^*$ & \textbf{1.080}$^*$ & \ColorDown{0.063} \textbf{1.137} \\
+ Conceptor-2 (extended) & \textbf{1.610}$^*$ & \textbf{0.479}$^*$ & \textbf{0.866}$^*$ & \textbf{1.119}$^*$ & \textbf{-0.215} & \textbf{0.727}$^*$ & \ColorDown{0.364} \textbf{0.836} \\
+ Conceptor-2 (propernouns) & 1.751$^*$ & 0.893$^*$ & \textbf{1.082}$^*$ & \textbf{1.125}$^*$ & \textbf{0.581}$^*$ & \textbf{1.167}$^*$ & \ColorDown{0.100} \textbf{1.100} \\
+ Conceptor-2 (all) & 1.773$^*$ & 0.862$^*$ & \textbf{1.101}$^*$ & \textbf{1.120}$^*$ & \textbf{0.628}$^*$ & \textbf{1.191}$^*$ & \ColorDown{0.088} \textbf{1.112} \\
+ Conceptor-2 (or) & 1.736$^*$ & \textbf{0.582}$^*$ & \textbf{0.702}$^*$ & \textbf{0.946}$^*$ & \textbf{-0.103} & \textbf{0.763}$^*$ & \ColorDown{0.395} \textbf{0.805} \\
\hdashline
(Wordlist Percentile 0.1)\\
+ Conceptor-2 (pronouns) & 1.828$^*$ & 0.971$^*$ & \textbf{1.185}$^*$ & \textbf{1.065}$^*$ & \textbf{0.755}$^*$ & \textbf{1.123}$^*$ & \ColorDown{0.046} \textbf{1.154} \\
+ Conceptor-2 (extended) & \textbf{1.638}$^*$ & \textbf{0.511}$^*$ & \textbf{1.167}$^*$ & \textbf{1.114}$^*$ & \textbf{0.265} & \textbf{1.017}$^*$ & \ColorDown{0.248} \textbf{0.952} \\
+ Conceptor-2 (propernouns) & 1.777$^*$ & 0.941$^*$ & \textbf{1.121}$^*$ & \textbf{1.169}$^*$ & 0.885$^*$ & 1.332$^*$ & \ColorUp{0.004} 1.204 \\
+ Conceptor-2 (all) & 1.795$^*$ & 0.952$^*$ & \textbf{1.070}$^*$ & \textbf{1.129}$^*$ & \textbf{0.785}$^*$ & \textbf{1.269}$^*$ & \ColorDown{0.033} \textbf{1.167} \\
+ Conceptor-2 (or) & \textbf{1.706}$^*$ & \textbf{0.726}$^*$ & \textbf{0.990}$^*$ & \textbf{0.972}$^*$ & \textbf{0.455}$^*$ & \textbf{0.978}$^*$ & \ColorDown{0.229} \textbf{0.971} \\

\hline
\end{tabular}
\caption{\label{tab:seat-bert-tiny-different-percentiles-on-sst} SEAT effect size of gender debising. 
The impact of \textit{different percentiles of wordlist} (using \textit{UMAP} clustering) on \textit{SST} Corpus, \textit{bert-tiny} models. The top-3 best performance is colored in orange.
}
\end{table*}

\begin{table*}[thb!]
\centering
\small 
\begin{tabular}{lllllllr}
\hline Model & SEAT-6 & SEAT-6b & SEAT-7 & SEAT-7b & SEAT-8 & SEAT-8b & Avg. Abs. \\
\hline BERT-T & 1.735$^*$ & 0.797$^*$ & 1.294$^*$ & 1.243$^*$ & 0.837$^*$ & 1.293$^*$ & 1.200 \\
\hdashline

(Wordlist Percentile 1.0)\\
+ Conceptor-2 (pronouns) & \textbf{1.676}$^*$ & \textbf{0.389}$^*$ & \textbf{1.218}$^*$ & \textbf{1.095}$^*$ & \textbf{0.557}$^*$ & \textbf{1.008}$^*$ & \ColorDown{0.210} \textbf{0.990} \\
+ Conceptor-2 (extended) & \textbf{1.578}$^*$ & \textbf{0.507}$^*$ & \textbf{1.248}$^*$ & \textbf{1.220}$^*$ & \textbf{0.656}$^*$ & 1.351$^*$ & \ColorDown{0.107} \textbf{1.093} \\
+ Conceptor-2 (propernouns) & \textbf{1.713}$^*$ & \textbf{0.776}$^*$ & \textbf{1.184}$^*$ & 1.315$^*$ & \textbf{0.538}$^*$ & \textbf{1.193}$^*$ & \ColorDown{0.080} \textbf{1.120} \\
+ Conceptor-2 (all) & \textbf{1.660}$^*$ & \textbf{0.379}$^*$ & \textbf{1.248}$^*$ & \textbf{1.185}$^*$ & \textbf{0.486}$^*$ & \textbf{1.125}$^*$ & \ColorDown{0.186} \textbf{1.014} \\
+ Conceptor-2 (or) & \textbf{1.550}$^*$ & \textbf{0.180} & \textbf{1.010}$^*$ & \textbf{1.146}$^*$ & \textbf{0.197} & \textbf{1.088}$^*$ & \ColorDown{0.338} \ColorBest{\textbf{0.862}} \\
\hdashline
(Wordlist Percentile 0.5-0.9)\\
+ Conceptor-2 (pronouns) & \textbf{1.676}$^*$ & \textbf{0.389}$^*$ & \textbf{1.218}$^*$ & \textbf{1.095}$^*$ & \textbf{0.557}$^*$ & \textbf{1.008}$^*$ & \ColorDown{0.210} \textbf{0.990} \\
+ Conceptor-2 (extended) & \textbf{1.684}$^*$ & \textbf{0.374} & \textbf{1.204}$^*$ & \textbf{1.065}$^*$ & \textbf{0.616}$^*$ & \textbf{1.144}$^*$ & \ColorDown{0.186} \textbf{1.014} \\
+ Conceptor-2 (propernouns) & \textbf{1.713}$^*$ & \textbf{0.776}$^*$ & \textbf{1.184}$^*$ & 1.315$^*$ & \textbf{0.538}$^*$ & \textbf{1.193}$^*$ & \ColorDown{0.080} \textbf{1.120} \\
+ Conceptor-2 (all) & \textbf{1.660}$^*$ & \textbf{0.379}$^*$ & \textbf{1.248}$^*$ & \textbf{1.185}$^*$ & \textbf{0.486}$^*$ & \textbf{1.125}$^*$ & \ColorDown{0.186} \textbf{1.014} \\
+ Conceptor-2 (or) & \textbf{1.573}$^*$ & \textbf{0.179} & \textbf{0.963}$^*$ & \textbf{1.117}$^*$ & \textbf{0.103} & \textbf{0.963}$^*$ & \ColorDown{0.384} \ColorBest{\textbf{0.816}} \\
\hdashline
(Wordlist Percentile 0.4)\\
+ Conceptor-2 (pronouns) & \textbf{1.677}$^*$ & \textbf{0.382}$^*$ & \textbf{1.218}$^*$ & \textbf{1.095}$^*$ & \textbf{0.561}$^*$ & \textbf{1.008}$^*$ & \ColorDown{0.210} \textbf{0.990} \\
+ Conceptor-2 (extended) & \textbf{1.578}$^*$ & \textbf{0.507}$^*$ & \textbf{1.248}$^*$ & \textbf{1.220}$^*$ & \textbf{0.656}$^*$ & 1.351$^*$ & \ColorDown{0.107} \textbf{1.093} \\
+ Conceptor-2 (propernouns) & \textbf{1.622}$^*$ & \textbf{0.745}$^*$ & \textbf{0.946}$^*$ & \textbf{0.937}$^*$ & \textbf{0.717}$^*$ & \textbf{1.104}$^*$ & \ColorDown{0.188} \textbf{1.012} \\
+ Conceptor-2 (all) & \textbf{1.656}$^*$ & \textbf{0.388}$^*$ & \textbf{1.198}$^*$ & \textbf{1.114}$^*$ & \textbf{0.490}$^*$ & \textbf{1.083}$^*$ & \ColorDown{0.212} \textbf{0.988} \\
+ Conceptor-2 (or) & \textbf{1.544}$^*$ & \textbf{0.270} & \textbf{0.899}$^*$ & \textbf{1.034}$^*$ & \textbf{0.497}$^*$ & \textbf{1.063}$^*$ & \ColorDown{0.316} \ColorBest{\textbf{0.884}} \\
\hdashline
(Wordlist Percentile 0.3)\\
+ Conceptor-2 (pronouns) & \textbf{1.554}$^*$ & \textbf{0.447}$^*$ & \textbf{1.116}$^*$ & \textbf{1.198}$^*$ & 0.973$^*$ & 1.429$^*$ & \ColorDown{0.080} \textbf{1.120} \\
+ Conceptor-2 (extended) & \textbf{1.584}$^*$ & \textbf{0.523}$^*$ & \textbf{1.248}$^*$ & \textbf{1.223}$^*$ & \textbf{0.636}$^*$ & 1.345$^*$ & \ColorDown{0.107} \textbf{1.093} \\
+ Conceptor-2 (propernouns) & \textbf{1.648}$^*$ & \textbf{0.764}$^*$ & \textbf{1.087}$^*$ & \textbf{1.134}$^*$ & 1.084$^*$ & 1.349$^*$ & \ColorDown{0.022} \textbf{1.178} \\
+ Conceptor-2 (all) & \textbf{1.588}$^*$ & \textbf{0.783}$^*$ & \textbf{1.125}$^*$ & \textbf{1.132}$^*$ & 0.989$^*$ & 1.325$^*$ & \ColorDown{0.043} \textbf{1.157} \\
+ Conceptor-2 (or) & \textbf{1.549}$^*$ & \textbf{0.402}$^*$ & \textbf{1.236}$^*$ & \textbf{1.186}$^*$ & 1.024$^*$ & 1.430$^*$ & \ColorDown{0.062} \textbf{1.138} \\
\hdashline
(Wordlist Percentile 0.2)\\
+ Conceptor-2 (pronouns) & \textbf{1.653}$^*$ & \textbf{0.588}$^*$ & \textbf{1.124}$^*$ & \textbf{1.084}$^*$ & 0.863$^*$ & \textbf{1.257}$^*$ & \ColorDown{0.105} \textbf{1.095} \\
+ Conceptor-2 (extended) & \textbf{1.584}$^*$ & \textbf{0.523}$^*$ & \textbf{1.248}$^*$ & \textbf{1.223}$^*$ & \textbf{0.636}$^*$ & 1.345$^*$ & \ColorDown{0.107} \textbf{1.093} \\
+ Conceptor-2 (propernouns) & \textbf{1.623}$^*$ & \textbf{0.753}$^*$ & \textbf{1.184}$^*$ & \textbf{1.224}$^*$ & 1.001$^*$ & 1.336$^*$ & \ColorDown{0.013} \textbf{1.187} \\
+ Conceptor-2 (all) & \textbf{1.600}$^*$ & \textbf{0.747}$^*$ & \textbf{1.190}$^*$ & \textbf{1.195}$^*$ & 0.986$^*$ & 1.314$^*$ & \ColorDown{0.028} \textbf{1.172} \\
+ Conceptor-2 (or) & \textbf{1.476}$^*$ & \textbf{0.364} & \textbf{1.089}$^*$ & \textbf{1.026}$^*$ & \textbf{0.407} & \textbf{1.162}$^*$ & \ColorDown{0.279} \textbf{0.921} \\
\hdashline
(Wordlist Percentile 0.1)\\
+ Conceptor-2 (pronouns) & \textbf{1.668}$^*$ & \textbf{0.775}$^*$ & \textbf{1.154}$^*$ & \textbf{0.989}$^*$ & \textbf{0.813}$^*$ & \textbf{1.171}$^*$ & \ColorDown{0.105} \textbf{1.095} \\
+ Conceptor-2 (extended) & \textbf{1.689}$^*$ & \textbf{0.762}$^*$ & 1.345$^*$ & \textbf{1.206}$^*$ & 0.991$^*$ & \textbf{1.260}$^*$ & \ColorUp{0.009} 1.209 \\
+ Conceptor-2 (propernouns) & \textbf{1.705}$^*$ & 0.863$^*$ & \textbf{1.257}$^*$ & \textbf{1.214}$^*$ & \textbf{0.812}$^*$ & \textbf{1.282}$^*$ & \ColorDown{0.011} \textbf{1.189} \\
+ Conceptor-2 (all) & \textbf{1.709}$^*$ & 0.870$^*$ & \textbf{1.224}$^*$ & \textbf{1.203}$^*$ & 0.858$^*$ & \textbf{1.287}$^*$ & \ColorDown{0.008} \textbf{1.192} \\
+ Conceptor-2 (or) & \textbf{1.630}$^*$ & \textbf{0.739}$^*$ & \textbf{1.069}$^*$ & \textbf{0.928}$^*$ & \textbf{0.753}$^*$ & \textbf{1.152}$^*$ & \ColorDown{0.155} \textbf{1.045} \\

\hline
\end{tabular}
\caption{\label{tab:seat-bert-tiny-different-percentiles-on-reddit} SEAT effect size of gender debising. 
The impact of \textit{different percentiles of wordlist} (using \textit{UMAP} clustering) on \textit{Reddit} Corpus, \textit{bert-tiny} models.  The top-3 best performance is colored in orange.
}
\end{table*}


\begin{table*}[thb!]
\centering
\small
\begin{tabular}{lllllllr}
\hline Model & SEAT-6 & SEAT-6b & SEAT-7 & SEAT-7b & SEAT-8 & SEAT-8b & Avg. Abs. \\
\hline
(Layer 0)\\
BERT-T & 1.536$^*$ & 0.640$^*$ & 0.959$^*$ & 1.307$^*$ & 0.263 & 0.814$^*$ & 0.920 \\
+ Conceptor-0 (or) & \textbf{0.803}$^*$ & \textbf{0.103} & \textbf{0.249} & \textbf{0.825}$^*$ & \textbf{0.039} & \textbf{0.568}$^*$ & \ColorDown{0.489} \textbf{0.431} \\
+ Conceptor-Intervened & \textbf{0.803}$^*$ & \textbf{0.103} & \textbf{0.249} & \textbf{0.825}$^*$ & \textbf{0.039} & \textbf{0.568}$^*$ & \ColorDown{0.489} \textbf{0.431} \\

\hdashline
(Layer 1)\\
BERT-T & 1.702$^*$ & 1.019$^*$ & 1.102$^*$ & 1.250$^*$ & 0.313 & 1.094$^*$ & 1.080 \\
+ Conceptor-1 (or) & \textbf{1.241}$^*$ & \textbf{-0.067} & \textbf{0.588}$^*$ & \textbf{0.939}$^*$ & -0.340 & \textbf{0.477}$^*$ & \ColorDown{0.471} \textbf{0.609} \\
+ Conceptor-Intervened & \textbf{0.928}$^*$ & \textbf{0.022} & \textbf{-0.427} & \textbf{0.708}$^*$ & -0.753 & \textbf{0.542}$^*$ & \ColorDown{0.517} \ColorBest{\textbf{0.563}} \\

\hdashline
(Layer 2)\\
BERT-T & 1.735$^*$ & 0.797$^*$ & 1.294$^*$ & 1.243$^*$ & 0.837$^*$ & 1.293$^*$ & 1.200 \\
+ Conceptor-2 (or) & \textbf{1.542}$^*$ & \textbf{0.148} & \textbf{0.486}$^*$ & \textbf{0.806}$^*$ & \textbf{-0.549} & \textbf{0.245} & \ColorDown{0.571} \textbf{0.629} \\
+ Conceptor-Intervened & \textbf{1.026}$^*$ & \textbf{-0.079} & \textbf{-0.264} & \textbf{0.862}$^*$ & \textbf{-0.500} & \textbf{0.239} & \ColorDown{0.705} \ColorBest{\textbf{0.495}} \\

\hline
\end{tabular}
\caption{\label{tab:seat-bert-tiny-different-layer-conceptor-intervened} SEAT effect size of gender debising from CI-BERT, Type I.
The conceptor-intervened performance of \textit{different layer's conceptor matrix} on \textit{SST} Corpus, \textit{bert-tiny} models. The layer(s) of CI-BERT that outperform the conceptor post-processing of the same layer(s) are colored in orange.
}
\end{table*}


\section{Full GPT2 Model Debiasing Results\label{app:additional-gpt2}}

\begin{itemize}
    \item Table~\ref{tab:gpt2-different-percentiles-on-brown} shows the post-processing gender debiasing result of different percentiles of wordlist on Brown corpus. 
\end{itemize}

\begin{table*}[thb!]
\centering
\small
\begin{tabular}{lllllllr}
\hline Model & SEAT-6 & SEAT-6b & SEAT-7 & SEAT-7b & SEAT-8 & SEAT-8b & Avg. Abs. \\
\hline 
GPT2 & -0.510 & 0.057 & -0.274 & -0.186 & -0.369 & -0.313 & 0.285 \\
\hdashline
(Wordlist Percentile 1.0)\\
+ Conceptor-12 (pronouns) & \textbf{0.030} & 0.269 & \textbf{-0.137} & \textbf{0.044} & \textbf{-0.129} & \textbf{-0.076} & \ColorDown{0.171} \ColorBest{\textbf{0.114}} \\
+ Conceptor-12 (extended) & \textbf{0.053} & 0.293 & \textbf{-0.163} & \textbf{0.057} & \textbf{-0.114} & \textbf{-0.054} & \ColorDown{0.163} \textbf{0.122} \\
+ Conceptor-12 (propernouns) & 0.713$^*$ & 0.430$^*$ & \textbf{0.029} & 0.222 & \textbf{0.045} & \textbf{0.196} & \ColorDown{0.013} \textbf{0.272} \\
+ Conceptor-12 (all) & \textbf{0.443}$^*$ & 0.333 & \textbf{-0.107} & \textbf{0.102} & \textbf{-0.088} & \textbf{0.065} & \ColorDown{0.095} \textbf{0.190} \\
+ Conceptor-12 (or) & \textbf{0.494}$^*$ & 0.269 & \textbf{0.065} & 0.236 & \textbf{-0.044} & \textbf{0.308} & \ColorDown{0.049} \textbf{0.236} \\
\hdashline
(Wordlist Percentile 0.9)\\
+ Conceptor-12 (pronouns) & \textbf{0.030} & 0.269 & \textbf{-0.137} & \textbf{0.044} & \textbf{-0.129} & \textbf{-0.076} & \ColorDown{0.171} \ColorBest{\textbf{0.114}} \\
+ Conceptor-12 (extended) & \textbf{0.312} & 0.448$^*$ & -0.352 & \textbf{0.032} & \textbf{-0.067} & \textbf{0.048} & \ColorDown{0.075} \textbf{0.210} \\
+ Conceptor-12 (propernouns) & 0.713$^*$ & 0.430$^*$ & \textbf{0.029} & 0.222 & \textbf{0.045} & \textbf{0.196} & \ColorDown{0.013} \textbf{0.272} \\
+ Conceptor-12 (all) & \textbf{0.443}$^*$ & 0.333 & \textbf{-0.107} & \textbf{0.102} & \textbf{-0.088} & \textbf{0.065} & \ColorDown{0.095} \textbf{0.190} \\
+ Conceptor-12 (or) & 0.537$^*$ & 0.268 & \textbf{0.051} & 0.260 & \textbf{-0.043} & 0.361 & \ColorDown{0.032} \textbf{0.253} \\
\hdashline
(Wordlist Percentile 0.8)\\
+ Conceptor-12 (pronouns) & \textbf{0.030} & 0.269 & \textbf{-0.137} & \textbf{0.044} & \textbf{-0.129} & \textbf{-0.076} & \ColorDown{0.171} \ColorBest{\textbf{0.114}} \\
+ Conceptor-12 (extended) & \textbf{0.312} & 0.448$^*$ & -0.352 & \textbf{0.032} & \textbf{-0.067} & \textbf{0.048} & \ColorDown{0.075} \textbf{0.210} \\
+ Conceptor-12 (propernouns) & 0.713$^*$ & 0.430$^*$ & \textbf{0.029} & 0.222 & \textbf{0.045} & \textbf{0.196} & \ColorDown{0.013} \textbf{0.272} \\
+ Conceptor-12 (all) & \textbf{0.443}$^*$ & 0.333 & \textbf{-0.107} & \textbf{0.102} & \textbf{-0.088} & \textbf{0.065} & \ColorDown{0.095} \textbf{0.190} \\
+ Conceptor-12 (or) & 0.537$^*$ & 0.268 & \textbf{0.051} & 0.260 & \textbf{-0.043} & 0.361 & \ColorDown{0.032} \textbf{0.253} \\
\hdashline
(Wordlist Percentile 0.7)\\
+ Conceptor-12 (pronouns) & \textbf{0.031} & 0.269 & \textbf{-0.136} & \textbf{0.045} & \textbf{-0.128} & \textbf{-0.076} & \ColorDown{0.171} \ColorBest{\textbf{0.114}} \\
+ Conceptor-12 (extended) & \textbf{0.312} & 0.448$^*$ & -0.352 & \textbf{0.032} & \textbf{-0.067} & \textbf{0.048} & \ColorDown{0.075} \textbf{0.210} \\
+ Conceptor-12 (propernouns) & 0.713$^*$ & 0.430$^*$ & \textbf{0.029} & 0.222 & \textbf{0.045} & \textbf{0.196} & \ColorDown{0.013} \textbf{0.272} \\
+ Conceptor-12 (all) & \textbf{0.443}$^*$ & 0.333 & \textbf{-0.107} & \textbf{0.102} & \textbf{-0.088} & \textbf{0.065} & \ColorDown{0.095} \textbf{0.190} \\
+ Conceptor-12 (or) & 0.538$^*$ & 0.269 & \textbf{0.051} & 0.260 & \textbf{-0.043} & 0.361 & \ColorDown{0.031} \textbf{0.254} \\
\hdashline
(Wordlist Percentile 0.6)\\
+ Conceptor-12 (pronouns) & \textbf{0.031} & 0.269 & \textbf{-0.136} & \textbf{0.045} & \textbf{-0.128} & \textbf{-0.076} & \ColorDown{0.171} \ColorBest{\textbf{0.114}} \\
+ Conceptor-12 (extended) & \textbf{0.304} & 0.449$^*$ & -0.381 & \textbf{-0.002} & \textbf{-0.106} & \textbf{0.016} & \ColorDown{0.075} \textbf{0.210} \\
+ Conceptor-12 (propernouns) & 0.713$^*$ & 0.430$^*$ & \textbf{0.029} & 0.222 & \textbf{0.045} & \textbf{0.196} & \ColorDown{0.013} \textbf{0.272} \\
+ Conceptor-12 (all) & \textbf{0.443}$^*$ & 0.333 & \textbf{-0.107} & \textbf{0.102} & \textbf{-0.088} & \textbf{0.065} & \ColorDown{0.095} \textbf{0.190} \\
+ Conceptor-12 (or) & 0.519$^*$ & 0.255 & \textbf{0.036} & 0.244 & \textbf{-0.054} & 0.348 & \ColorDown{0.042} \textbf{0.243} \\
\hdashline
(Wordlist Percentile 0.5)\\
+ Conceptor-12 (pronouns) & \textbf{0.031} & 0.269 & \textbf{-0.136} & \textbf{0.045} & \textbf{-0.128} & \textbf{-0.076} & \ColorDown{0.171} \ColorBest{\textbf{0.114}} \\
+ Conceptor-12 (extended) & \textbf{0.269} & 0.411$^*$ & \textbf{-0.258} & \textbf{-0.038} & \textbf{0.114} & \textbf{0.027} & \ColorDown{0.099} \textbf{0.186} \\
+ Conceptor-12 (propernouns) & 0.713$^*$ & 0.430$^*$ & \textbf{0.029} & 0.222 & \textbf{0.045} & \textbf{0.196} & \ColorDown{0.013} \textbf{0.272} \\
+ Conceptor-12 (all) & 0.565$^*$ & 0.445$^*$ & \textbf{-0.110} & \textbf{0.125} & \textbf{-0.063} & \textbf{0.127} & \ColorDown{0.046} \textbf{0.239} \\
+ Conceptor-12 (or) & \textbf{0.478}$^*$ & 0.248 & \textbf{0.029} & 0.188 & \textbf{-0.016} & 0.317 & \ColorDown{0.072} \textbf{0.213} \\
\hdashline
(Wordlist Percentile 0.4)\\
+ Conceptor-12 (pronouns) & \textbf{0.060} & 0.271 & \textbf{-0.130} & \textbf{0.045} & \textbf{-0.131} & \textbf{-0.074} & \ColorDown{0.167} \textbf{0.118} \\
+ Conceptor-12 (extended) & \textbf{0.276} & 0.410$^*$ & \textbf{-0.255} & \textbf{-0.038} & \textbf{0.110} & \textbf{0.027} & \ColorDown{0.099} \textbf{0.186} \\
+ Conceptor-12 (propernouns) & 0.730$^*$ & 0.393$^*$ & \textbf{0.033} & 0.190 & \textbf{0.028} & \textbf{0.127} & \ColorDown{0.035} \textbf{0.250} \\
+ Conceptor-12 (all) & 0.648$^*$ & 0.374$^*$ & \textbf{-0.036} & \textbf{0.154} & \textbf{0.014} & \textbf{0.134} & \ColorDown{0.058} \textbf{0.227} \\
+ Conceptor-12 (or) & 0.564$^*$ & 0.258 & \textbf{0.057} & 0.196 & \textbf{0.014} & \textbf{0.287} & \ColorDown{0.056} \textbf{0.229} \\
\hdashline
(Wordlist Percentile 0.3)\\
+ Conceptor-12 (pronouns) & \textbf{0.092} & 0.316 & \textbf{-0.001} & \textbf{0.064} & \textbf{-0.035} & \textbf{-0.062} & \ColorDown{0.190} \ColorBest{\textbf{0.095}} \\
+ Conceptor-12 (extended) & \textbf{0.264} & 0.369 & \textbf{-0.261} & \textbf{-0.043} & \textbf{0.115} & \textbf{0.015} & \ColorDown{0.107} \textbf{0.178} \\
+ Conceptor-12 (propernouns) & 0.729$^*$ & 0.391$^*$ & \textbf{0.030} & 0.187 & \textbf{0.028} & \textbf{0.125} & \ColorDown{0.037} \textbf{0.248} \\
+ Conceptor-12 (all) & 0.682$^*$ & 0.397$^*$ & \textbf{-0.021} & \textbf{0.162} & \textbf{0.018} & \textbf{0.140} & \ColorDown{0.048} \textbf{0.237} \\
+ Conceptor-12 (or) & 0.545$^*$ & 0.214 & \textbf{0.072} & 0.198 & \textbf{0.006} & \textbf{0.279} & \ColorDown{0.066} \textbf{0.219} \\
\hdashline
(Wordlist Percentile 0.2)\\
+ Conceptor-12 (pronouns) & \textbf{0.094} & 0.316 & \textbf{-0.001} & \textbf{0.064} & \textbf{-0.033} & \textbf{-0.062} & \ColorDown{0.190} \ColorBest{\textbf{0.095}} \\
+ Conceptor-12 (extended) & \textbf{0.249} & 0.356 & \textbf{-0.238} & \textbf{-0.048} & \textbf{0.106} & \textbf{0.016} & \ColorDown{0.116} \textbf{0.169} \\
+ Conceptor-12 (propernouns) & 0.699$^*$ & 0.403$^*$ & \textbf{0.042} & \textbf{0.176} & \textbf{0.017} & \textbf{0.092} & \ColorDown{0.047} \textbf{0.238} \\
+ Conceptor-12 (all) & 0.699$^*$ & 0.437$^*$ & \textbf{0.056} & \textbf{0.164} & \textbf{0.063} & \textbf{0.098} & \ColorDown{0.032} \textbf{0.253} \\
+ Conceptor-12 (or) & 0.519$^*$ & 0.223 & \textbf{0.086} & 0.191 & \textbf{-0.004} & \textbf{0.255} & \ColorDown{0.072} \textbf{0.213} \\
\hdashline
(Wordlist Percentile 0.1)\\
+ Conceptor-12 (pronouns) & \textbf{0.446}$^*$ & 0.599$^*$ & \textbf{0.035} & \textbf{0.064} & \textbf{0.064} & \textbf{-0.037} & \ColorDown{0.077} \textbf{0.208} \\
+ Conceptor-12 (extended) & \textbf{0.497}$^*$ & 0.544$^*$ & \textbf{0.019} & \textbf{0.097} & \textbf{0.171} & \textbf{0.074} & \ColorDown{0.051} \textbf{0.234} \\
+ Conceptor-12 (propernouns) & 0.753$^*$ & 0.495$^*$ & \textbf{-0.038} & \textbf{0.095} & \textbf{-0.010} & \textbf{0.062} & \ColorDown{0.043} \textbf{0.242} \\
+ Conceptor-12 (all) & 0.730$^*$ & 0.508$^*$ & \textbf{-0.009} & \textbf{0.086} & \textbf{0.102} & \textbf{0.070} & \ColorDown{0.034} \textbf{0.251} \\
+ Conceptor-12 (or) & 0.914$^*$ & 0.568$^*$ & \textbf{0.258} & 0.247 & \textbf{0.021} & \textbf{0.264} & \ColorUp{0.094} \textbf{0.379} \\
\hline
\end{tabular}
\caption{\label{tab:gpt2-different-percentiles-on-brown} SEAT effect size of gender debising. The impact of \textit{different percentiles of wordlist} (using \textit{UMAP} clustering) on \textit{Brown} Corpus, \textit{gpt-2} models. The top-3 best performance is colored in orange.}
\end{table*}


\section{Full Other LLMs' GLUE Results\label{app:additional-llms}}

\begin{itemize}
    \item Table~\ref{tab:glue-llm-full} contains GLUE results for gender debiased model. 
\end{itemize}

\begin{table*}[]
    \centering
    \small
    \begin{tabular}{lrrrrrrrrrrr}
\hline Model & CoLA & MNLI & MRPC & QNLI & QQP & RTE & SST & STS-B & WNLI & Average \\
\hline
BERT-L & 62.82 & 86.13 & 88.32 & 92.15 & 91.56 & 69.31 & 93.81 & 90.00 & 35.21 & 78.81 \\
+ Conceptor P.P. & 62.34 & 86.16 & 89.44 & 91.71 & 91.59 & 74.73 & 93.58 & 90.06 & 30.17 & \ColorUpCR{0.05} 78.86 \\
\hline 
GPT2 & 29.10 & 82.43 & 84.51 & 87.71 & 89.18 & 64.74 & 91.97 & 84.26 & 43.19 & 73.01 \\
+ Conceptor P.P. & 35.47 & 82.39 & 84.08 & 88.30 & 89.12 & 67.15 & 92.09 & 83.67 & 33.80 & \ColorDownCR{0.11} 72.90 \\
\hline 
GPT2-L & 12.48 & 88.80 & 82.98 & 80.30 & 87.61 & 51.26 & 81.77 & 78.87 & 46.48 & 75.84 \\
+ Conceptor P.P. & 20.03 & 88.92 & 82.78 & 79.64 & 87.65 & 50.54 & 82.34 & 78.26 & 40.85 &  \ColorUpCR{0.04} 75.89 \\
\hline
GPT-J & 59.73 & 82.49 & 87.95 & 87.93 & 87.54 & 75.81 & 94.50 & 88.60 & 39.44 & 78.22 \\
+ Conceptor & 59.48 & 82.89 & 87.69 & 91.56 & 89.90 & 76.17 & 95.07 & 88.81 & 30.99 &    \ColorDownCR{0.14} 78.06 \\
\hline
\end{tabular}
    \caption{GLUE validation set results for other LLMs. We use the F1 score for MRPC, the Spearman correlation for STS-B,and Matthew's correlation for CoLA. For all other tasks,we report the accuracy. }
    \label{tab:glue-llm-full}
\end{table*}


\section{Full Intersectional Debiasing Results\label{app:intersect}}

\begin{itemize}
    \item Table~\ref{tab:intersect-gender} and~\ref{tab:intersect-race} show the post-processing intersectional debiasing results. 
\end{itemize}

\begin{table*}[thb!]
\centering
\small
\begin{tabular}{lllllllr}
\hline Model & SEAT-6 & SEAT-6b & SEAT-7 & SEAT-7b & SEAT-8 & SEAT-8b & Avg. Abs. \\
\hline 
BERT & 0.931$^*$ & 0.090 & -0.124 & 0.937$^*$ & 0.783$^*$ & 0.858$^*$ & 0.620 \\
+ Gender Conceptor & \textbf{0.388} & \textbf{-0.078} & -0.292 & \textbf{0.179} & \textbf{0.594}$^*$ & \textbf{0.335} & \ColorDown{0.309} \textbf{0.311} \\
+ Intersected Conceptor & \textbf{0.916}$^*$ & \textbf{0.026} & -0.180 & \textbf{0.840}$^*$ & \textbf{0.749}$^*$ & \textbf{0.832}$^*$ & \ColorDown{0.029} \textbf{0.591} \\
\hline
\end{tabular}
\caption{\label{tab:intersect-gender} BERT intersectional gender debiasing, where intersected conceptor indicates the conceptor matrix generated by its negated AND operation of gender conceptor matrix and race conceptor matrix
}
\end{table*}

\begin{table*}[thb!]
\centering
\small
\begin{tabular}{llllllllr}
\hline Model & ABW-1 & ABW-2 & SEAT-3 & SEAT-3b & SEAT-4  & SEAT-5 & SEAT-5b & Avg. Abs. \\
\hline 
BERT & -0.079 & 0.690$^*$ & 0.778$^*$ & 0.469$^*$ & 0.901$^*$ & 0.887$^*$ & 0.539$^*$ & 0.620 \\
+ Race Conceptor & \textbf{-0.063} & \textbf{0.682}$^*$ & 0.803$^*$ & \textbf{0.209} & 0.949$^*$ & 0.946$^*$ & \textbf{0.390}$^*$ & \ColorDown{0.043} \textbf{0.577} \\
+ Intersected Conceptor & \textbf{-0.045} & \textbf{0.685}$^*$ & 0.799$^*$ & \textbf{0.361}$^*$ & 0.926$^*$ & 0.931$^*$ & \textbf{0.484}$^*$ & \ColorDown{0.016} \textbf{0.604} \\
\hline
\end{tabular}
\caption{\label{tab:intersect-race} BERT intersectional race debiasing, where intersected conceptor indicates the conceptor matrix generated by its negated AND operation of gender conceptor matrix and race conceptor matrix
}
\end{table*}

\end{document}